\pdfoutput=1
\documentclass[11pt]{article}

\usepackage{acl}

\usepackage{times}
\usepackage{latexsym}

\usepackage[T1]{fontenc}
\usepackage[utf8]{inputenc}

\usepackage{microtype}

\usepackage{inconsolata}

\usepackage{graphicx}

\usepackage{hyperref}
\usepackage{url}

\usepackage{xcolor}
\usepackage{graphicx}
\usepackage{multirow}
\usepackage{makecell}
\usepackage{pifont}       
\usepackage{bbding}      
\usepackage{fontawesome}
\usepackage{subcaption}
\usepackage{tcolorbox}
\usepackage{booktabs}
\usepackage{amsmath,amsfonts,bm}

\title{Graph of Records: Boosting Retrieval Augmented Generation for Long-context Summarization with Graphs}

\author{Haozhen Zhang$^{1}$\thanks{Work done as an intern at University of Illinois at Urbana-Champaign} \and Tao Feng \and Jiaxuan You \\
University of Illinois at Urbana-Champaign \\
\texttt{\{haozhenz,taofeng2,jiaxuan\}@illinois.edu, $^{1}$wazhz14@gmail.com} \\
}

\begin{document}
\maketitle
\begin{abstract}
Retrieval-augmented generation (RAG) has revitalized Large Language Models (LLMs) by injecting non-parametric factual knowledge. 
Compared with long-context LLMs, RAG is considered an effective summarization tool in a more concise and lightweight manner, which can interact with LLMs multiple times using diverse queries to get comprehensive responses. 
However, the LLM-generated historical responses, which contain potentially insightful information, are largely neglected and discarded by existing approaches, leading to suboptimal results. 
In this paper, we propose \textit{graph of records} (\textbf{GoR}), which leverages historical responses generated by LLMs to enhance RAG for long-context global summarization. 
Inspired by the \textit{retrieve-then-generate} paradigm of RAG, we construct a graph by establishing an edge between the retrieved text chunks and the corresponding LLM-generated response. 
To further uncover the intricate correlations between them, GoR features a \textit{graph neural network} and an elaborately designed \textit{BERTScore}-based objective for self-supervised model training, enabling seamless supervision signal backpropagation between reference summaries and node embeddings. 
We comprehensively compare GoR with 12 baselines across four long-context summarization datasets, and the results indicate that our proposed method reaches the best performance (\textit{e.g.}, 15\%, 8\%, and 19\% improvement over retrievers w.r.t. Rouge-L, Rouge-1, and Rouge-2 on the WCEP dataset). 
Extensive experiments further demonstrate the effectiveness of GoR. 
Code is available at \url{https://github.com/ulab-uiuc/GoR}
\end{abstract}

\section{Introduction}

Large Language Models (LLMs) have recently achieved remarkable performance across sorts of language modeling tasks~\citep{achiam2023GPT-4, llama3modelcard}. 
Among them, the long-context global summarization task is of great importance, which requires ultra-long context understanding capabilities of LLMs~\citep{li2024long-context-struggle, liu2024lost-middle}. 
Current attempts to accomplish this task mainly include long-context LLMs~\citep{touvron2023llama-2, glm2024chatglm, longchat2023, longllama} and retrieval-augmented generation (RAG)~\citep{RALM, yu2023ralm, trivedi2022interleaving-ralm, jiang2023active-ralm, asai2023self-rag}. 
In comparison with long-context LLMs that expand their context window to accommodate long-context inputs, RAG performs a cost-effective \textit{retrieve-then-generate} paradigm and provides a few retrieved short text chunks from a long document to LLMs. 
In a running RAG system, there are usually a large number of historical user queries and LLM-generated responses for a long document. 
Nevertheless, these historical responses, which contain informative task-related content, are mostly neglected without sufficient utilization by current RAG approaches.

Unfortunately, utilizing LLM historical responses for long-context global summarization presents two major challenges. 
(1) \textit{Sophisticated yet implicit correlations between historical responses and text}. 
Given a long document, there will inevitably be complicated correlations among plentiful user queries (\textit{e.g.}, logical correlations), which are further inherited by LLM-generated responses and the retrieved text chunks. 
However, uncovering these correlations is non-trivial since most text embeddings from language models (\textit{e.g.}, SBERT~\citep{Sentence-BERT}) or retrievers~\citep{karpukhin2020dense-dpr} concentrate on semantic similarity, which faces degrading performance in this case. 
(2) \textit{Lack of supervision signal}. 
In contrast with local (\textit{e.g.}, query-based) summarization~\citep{QMSum, wang2022squality} that includes golden reference text as labels, global summarization needs to be considered from the perspective of the long document as a whole and only has global reference summaries, which complicates the direct backpropagation of effective, accurate, and deterministic supervision signals to optimize the model towards a few relevant text chunks.

Based on the above observations, we propose \textit{graph of records} (\textbf{GoR}), which utilizes and organizes LLM historical responses as a graph of records for enhancing long-context global summarization in RAG. 
In detail, we first leverage LLMs to simulate some user queries conditioned on arbitrary text chunks in a long document to obtain historical responses under the paradigm of RAG, and an edge is then created between the retrieved text chunks and the LLM-generated response to construct a graph of records. 
To learn fine-grained correlations among nodes, we employ a \textit{graph neural network} and reuse the simulated user queries with the corresponding source text chunk as self-supervised training data. 
Intuitively, we hope the node embeddings can be adaptively learned to reflect the semantic and logical correlations with a given query. 
Inspired by the well-received BERTScore~\citep{zhang2019bertscore} that quantifies the semantic similarity between two paragraphs of text, we rely on it to rank the nodes according to their similarity with the self-supervised label of a given simulated query. 
In this way, node embeddings can benefit the indirect supervision signal from the self-supervised labels and be flexibly optimized using a contrastive loss and a pair-wise ranking loss based on the node rankings. 
In the experiments, we adopt four long-context summarization datasets, and the results demonstrate the superiority and effectiveness of our proposed method. 
For example, we show that GoR outperforms retrievers by 15\%, 8\%, and 19\% w.r.t. Rouge-L, Rouge-1, and Rouge-2, respectively, on the WCEP dataset. 
We also provide detailed comparisons and insightful analyses through extensive experiments, further showcasing the effectiveness of our approach.
Our contributions are summarized as follows:
\begin{itemize}
\item We propose \textit{graph of records} (\textbf{GoR}), which utilizes and organizes LLM-generated historical responses as a graph of records to strengthen RAG for long-context global summarization. We reveal that the fine-grained correlations between LLM historical responses and text chunks from long documents can be uncovered and utilized effectively to improve RAG performance. 
\item We leverage a ~\textit{graph neural network} and design a ~\textit{BERTScore}-based objective to optimize node embeddings, which can be adaptively learned in a self-supervised manner to reflect the semantic and complex correlations with input queries. Furthermore, the indirect supervision signal from self-supervised labels is crucial and conducive to the effective optimization of node embeddings. 
\item We evaluate our proposed method on four long-context summarization datasets, and the results show that GoR outperforms several competitive baselines by a significant margin. Extensive experiments and detailed analysis verify the superiority of GoR. 
\end{itemize}

\section{Graph of Records}

In this section, we first present some necessary backgrounds in Section~\ref{sec:pre}. 
Then, we describe our proposed method sequentially through three sections, \textit{i.e.}, Graph Construction (Section~\ref{method:graph}), BERTScore-based Objective for Self-supervised Training (Section~\ref{method:training}), and Retrieval from the Graph for Summarization (Section~\ref{method:retrieval}).

\subsection{Preliminaries}
\label{sec:pre}

\textbf{Retrieval-augmented Generation}. 
Retrieval-augmented Generation (RAG)~\citep{RALM} can typically be summarized into the following two processes. 
(1) \textit{Retrieval}. 
Give a long document which consists of several split text chunks $\mathbf{C} = \{\mathbf{c}_{\mathbf{i}}\}_{i=1}^{|\mathbf{C}|}$ as retrieval corpus, RAG first employs a retriever (\textit{e.g.}, Contriever~\citep{izacard2021unsupervised-contriever}) to retrieve $\mathbf{K}$ text chunks that are most relevant to a given query $\mathbf{q}$ based on semantic similarity. 
The retriever typically embeds the query $\mathbf{q}$ and a text chunk $\mathbf{c}$ from $\mathbf{C}$ using a query encoder $\mathbf{E}_{\mathbf{q}}(\cdot)$ and a context encoder $\mathbf{E}_{\mathbf{c}}(\cdot)$, respectively, and quantify their semantic similarity by the dot product operation described as $\operatorname{Sim}(\mathbf{q}, \mathbf{c}) = \mathbf{E}_{\mathbf{q}}(\mathbf{q})^{T} \cdot \mathbf{E}_{\mathbf{c}}(\mathbf{c})$. 
(2) \textit{Generation}. 
The retrieved text chunks are fed into LLMs with the query $\mathbf{q}$ to obtain the final response $\mathbf{r}$. 
The whole process can be described as:
\begin{equation}
\begin{split}
\mathbf{r} &= \operatorname{Generation}(\mathbf{q}, \{\mathbf{c}_{\mathbf{1}}, \cdots, \mathbf{c}_{\mathbf{K}}\}), \\ 
\{\mathbf{c}_{\mathbf{1}}, \cdots, \mathbf{c}_{\mathbf{K}}\} &= \operatorname{Retrieval}(\mathbf{q} | \mathbf{C}).
\end{split}
\end{equation}

\textbf{Graph Neural Networks}. 
Graph Neural Networks (GNNs)~\citep{GCN} stand out for their excellent representation learning ability on graph data. 
GNNs update node embeddings iteratively by aggregating messages from their neighboring nodes. 
Generally, the $l$-th layer of GNNs can be formalized as:
\begin{equation}
\begin{split}
\mathbf{h}_{v}^{(l)} = \mathrm{AGG}^{(l)}\Big(\mathbf{h}_{v}^{(l-1)}, \mathrm{MSG}^{(l)}\Big(\{\mathbf{h}_{u}^{(l-1)}, \Big.\Big. \\ u \in N(v)\}; \theta_m^l \Big.\Big); \theta_a^l \Big.\Big).
\end{split}
\end{equation}
where $\mathbf{h}_{u}^{(l)} \in \mathbb{R}^{d_{l}}$ is the embedding vector of nodes $u$ in layer $l$ and the dimension is $d_l$. 
$\mathrm{MSG}^{(l)}(\cdot)$ is a message computation function parameterized by $\theta_m^l$ and 
$\mathrm{AGG}^{(l)}(\cdot)$ is a message aggregation function parameterized by $\theta_a^l$ in layer $l$.

\subsection{Graph Construction}
\label{method:graph}

In this section, we describe how to organize LLM historical responses into a graph of records by simulating user queries. 

\textbf{Query Simulation.} User queries play a very critical role in the design of GoR since LLM historical responses generated by lots of repetitive, nonsense, or meaningless questions are inherently not beneficial for summarization. 
One solution is to use doc2query~\citep{nogueira2019doc2query} to simulate queries for a long document, but the generated results inevitably suffer from simplicity and rigidity due to the limited text generation capabilities of T5~\citep{T5}. 
To this end, we directly turn to LLMs for query simulation with temperature sampling instead of greedy decoding for generating meaningful, insightful, and diverse questions. 
Specifically, we split a long document into several text chunks $\mathbf{C}$ following the standard procedure of RAG and prompt LLMs to generate a query $\mathbf{q}^{\mathbf{s}}$ based on a randomly selected text chunk $\mathbf{c}^{\mathbf{s}}$. 
We repeat the above process until a certain number of non-duplicate queries are generated, which are gathered in pairs with the corresponding text chunks to form a corpus $\mathbf{T} = \{(\mathbf{q}^{\mathbf{s}}_{i}, \mathbf{c}^{\mathbf{s}}_{i})\}_{i=1}^{|\mathbf{T}|}$ for further model training (Section~\ref{method:training}). 
% The prompt for query simulation can be found in Appendix~\ref{appendix:prompts}. 

\textbf{Organize LLM Historical Responses into A Graph}. After obtaining simulated queries, we utilize them to perform RAG on the long document. 
LLM-generated responses during this process include informative and valuable understanding, summarizing, and answering of retrieved text chunks in the long document. 
Moreover, since there may exist sophisticated correlations among simulated queries, the text chunks and responses can inherit these features and potentially assist in answering a more comprehensive query, especially global summarization that needs to be understood from a holistic perspective. 
Nevertheless, it is a significant challenge to find correlations among complex and massive text at the linguistic level and the embeddings from language models (\textit{e.g.}, SBERT~\citep{Sentence-BERT}) or retrievers~\citep{karpukhin2020dense-dpr} focus on semantic similarity, which also suffers from poor performance in this case. 
To this end, we propose to break out of this dilemma by organizing these historical responses into a graph. 

Inspired by the \textit{retrieve-then-generate} process of RAG, we can connect the retrieved chunks to the corresponding response generated by LLMs since they are naturally relevant in content. 
Sequentially, during the $\mathbf{i}$-th round RAG, given the simulated query $\mathbf{q}^{\mathbf{s}}_{\mathbf{i}}$, we expand the retrieval corpus $\mathbf{C}$ with previously generated responses $\{\mathbf{r}_{\mathbf{1}}, \cdots, \mathbf{r}_{\mathbf{i-1}}\}$ and then build an edge between each retrieved chunk $\mathbf{c}_{\mathbf{j}} \in \{\mathbf{c}_{\mathbf{1}}, \cdots, \mathbf{c}_{\mathbf{K}}\}$ and the newly generated LLM response $\mathbf{r}_{\mathbf{i}}$, resulting in $\mathbf{K}$ edges constructed in each round. 
Note that we append the responses generated by each round of RAG to the retrieval corpus because they contain more refined knowledge compared with the text chunks from $\mathbf{C}$ and can help LLMs generate comprehensive responses in a self-evolving manner. 
Formally, the $\mathbf{i}$-th round RAG on simulated queries $\{\mathbf{q}^{\mathbf{s}}_{i}\}_{i=1}^{|\mathbf{T}|}$ can be described as:
\begin{equation}
\begin{split}
\mathbf{r}_{\mathbf{i}} \! &= \! \operatorname{Generation}(\mathbf{q}^{\mathbf{s}}_{\mathbf{i}}, \! \{\mathbf{c}_{\mathbf{1}}, \cdots, \mathbf{c}_{\mathbf{K}}\}), \\ 
\{\mathbf{c}_{\mathbf{1}}, \cdots, \mathbf{c}_{\mathbf{K}}\} \! &= \! \operatorname{Retrieval}(\mathbf{q}^{\mathbf{s}}_{\mathbf{i}} | \mathbf{C},\! \{\mathbf{r}_{\mathbf{1}}, \cdots, \mathbf{r}_{\mathbf{i-1}}\}).
\end{split}
\end{equation}

In this way, the LLM-generated responses serve as \textit{bridges} to connect the originally scattered text chunks $\mathbf{C}$ so that the fine-grained and sophisticated correlations among them can be better modeled and explored. 
Furthermore, we can potentially leverage historical responses generated by LLMs and enhance the quality of future LLM responses.

\begin{figure*}[t]
\begin{center}
        \centering
	\includegraphics[width=1.0\linewidth]{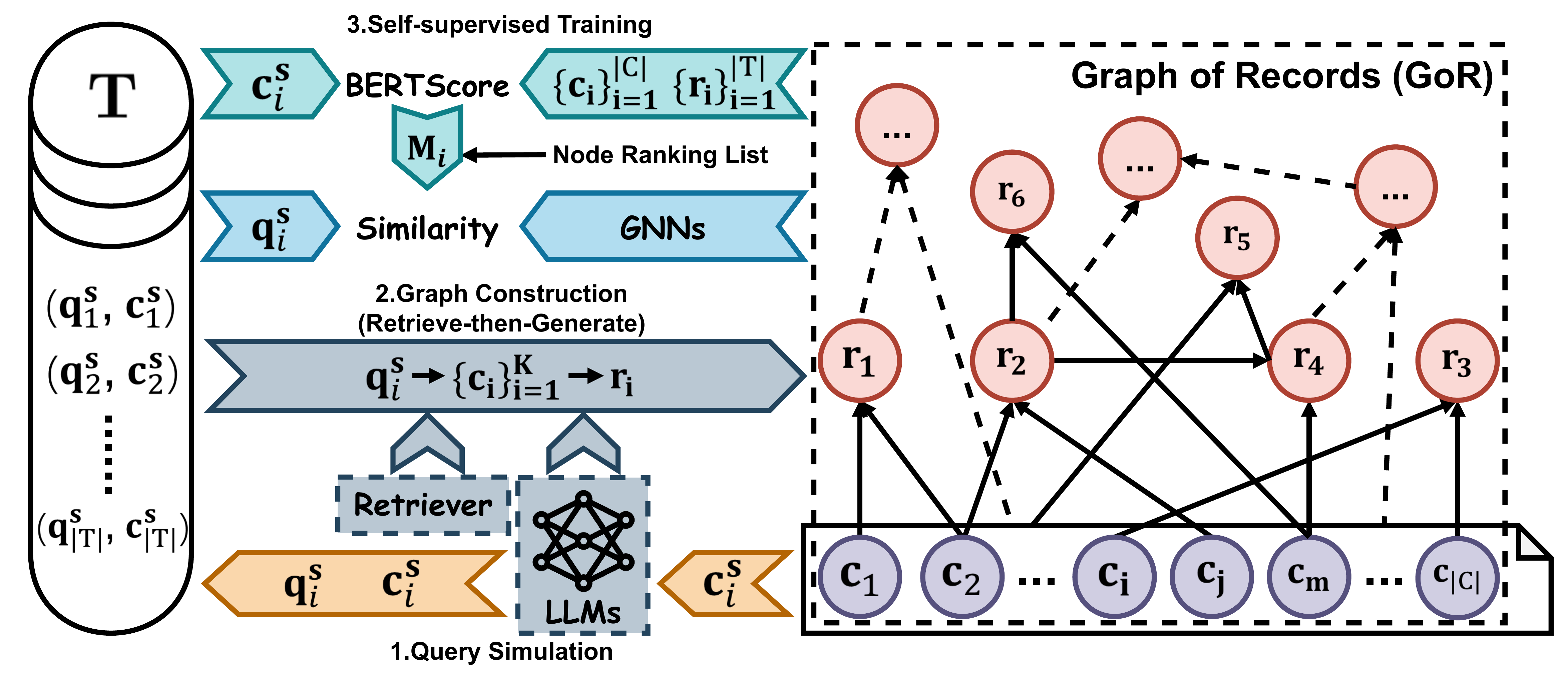}
	% \vspace{-5ex}
        \caption{\textbf{GoR model architecture}. GoR randomly selects text chunks $\mathbf{c}_{\mathbf{i}}$ from long documents to feed into LLMs for query simulation, which are saved as a self-supervised training corpus $\mathbf{T}$ and further used for graph construction inspired by the retrieve-then-generate paradigm in RAG. For model training, GoR leverages GNNs to obtain node embeddings and calculate their similarities to the query embedding. Finally, GoR features contrastive learning and pair-wise ranking objectives based on the node ranking list $\mathbf{M}_{i}$ derived from BERTScore calculation.}
        % \vspace{-1ex}
	\label{model_figure}
\end{center}
\end{figure*}

\subsection{BERTScore-based Objective for Self-supervised Training}
\label{method:training}

So far, we have constructed a graph using LLM-generated historical responses during RAG given the simulated queries. 
The key in this section lies in designing a reasonable and effective objective function for model optimization. 
Considering that some random walk~\citep{grover2016node2vec} or propagation-based~\citep{LabelPropagation} algorithms are not differentiable, we turn to graph neural networks (GNNs) for learning node embeddings, which are backpropagation-friendly. 
Intuitively, given a global summarization query $\mathbf{q}$, our ultimate optimization goal is to make the learned node embeddings adaptively reflect the similarity with the query embedding $\mathbf{E}_{\mathbf{q}}(\mathbf{q})$ by taking the complicated correlations among nodes into account. 
However, in global summarization tasks, there are essentially no text chunk indices as labels to indicate which nodes are most relevant for a query since it needs to consider the long document as a whole. 
Another naive solution is to use global reference summaries as labels, but there is a gap in supervision signal backpropagation between them and node embeddings because we still need to find out which nodes are most relevant to them.

Therefore, inspired by BERTScore~\citep{zhang2019bertscore}, which measures the semantic similarity between the reference and the generated text, we propose to use it to rank all nodes based on the similarity with reference summaries. 
By this means, BERTScore fills the gap in the backpropagation so that node embeddings can benefit the indirect supervision signal from the reference summaries. 
Nevertheless, global reference summaries contain broad information about long documents, making them highly semantically relevant to many nodes, which will confuse the model optimization direction and degrade the performance (we will discuss it in Section~\ref{sec:exp_sup}).

\textbf{Contrastive Loss Driven by BERTScore}. Based on the above observations, we directly reuse the simulated queries $\mathbf{T} = \{(\mathbf{q}^{\mathbf{s}}_{i}, \mathbf{c}^{\mathbf{s}}_{i})\}_{i=1}^{|\mathbf{T}|}$ to serve as self-supervised training data, in which the text chunk $\mathbf{c}^{\mathbf{s}}_{i}$ is highly relevant to the query $\mathbf{q}^{\mathbf{s}}_{i}$ and has more focused content. 
Given node embeddings output by the last $L$-th layer of GNNs, for the $i$-th query $\mathbf{q}^{\mathbf{s}}_{i}$, we rank them according to the similarity with the $i$-th text chunk $\mathbf{c}^{\mathbf{s}}_{i}$ and obtain a node embedding ranking list $\mathbf{M}_{i}$: 
\begin{equation}
\mathbf{M}_{i} = [\mathbf{h}_{+}^{(L)}, \mathbf{h}_{1}^{(L)}, \cdots, \mathbf{h}_{|\mathbf{C}| + |\mathbf{T}|}^{(L)}],
\end{equation}
where $\mathbf{h}_{+}^{(L)}$ stands for the node embedding with highest similarity. Note that we utilize the context encoder $\mathbf{E}_{\mathbf{c}}(\cdot)$ from the retriever to initialize node embeddings for simplicity. 
Then, we regard $\mathbf{h}_{+}^{(L)}$ as the positive while the rest in $\mathbf{M}_{i}$ as negative samples to conduct contrastive learning using InfoNCE~\citep{CPC}, which brings the query $\mathbf{q}^{\mathbf{s}}_{i}$ and the positive sample $\mathbf{h}_{+}^{(L)}$ closer in the semantic embedding space. 
We formulate the contrastive training objective as follows:
\begin{equation}
s(\mathbf{q}, \mathbf{h}) = \exp \left( {\mathbf{E}_{\mathbf{q}}(\mathbf{q})}^{\top} \mathbf{h} / \tau \right),
\end{equation}
\begin{equation}
\label{formula:cl}
\mathcal{L}_{\mathrm{CL}}\! =\! - \frac{1}{|\mathbf{T}|} \!\sum\limits_{j=1}^{|\mathbf{T}|} \log \! \frac{s(\mathbf{q}^{\mathbf{s}}_{j}, \mathbf{h}_{+}^{(L)})}{s(\mathbf{q}^{\mathbf{s}}_{j}, \mathbf{h}_{+}^{(L)}) \! + \!\!\!\!\!\sum\limits_{i=1}^{|\mathbf{M}_{j}-1|} \!\!\!\! s(\mathbf{q}^{\mathbf{s}}_{j}, \mathbf{h}_i^{(L)})},
\end{equation}
where $\tau$ is the temperature coefficient. 
Note that in the optimization pipeline of GoR, we conduct mini-batch training on the graph level, and each graph is associated with an independent self-supervised training dataset $\mathbf{T}$. 
We also leverage in-batch negatives from other graphs since the nodes in them are completely irrelevant content from other long documents (it is not shown in Formula~\ref{formula:cl} for brevity).

\textbf{Auxiliary Pair-wise Ranking Loss}. 
In the above-described contrastive loss $\mathcal{L}_{\mathrm{CL}}$, although we impose constraints on positive and negative samples, the ranking of negative samples themselves is not well utilized. 
Inspired by LambdaRank~\citep{lambdarank}, we further introduce an auxiliary pair-wise ranking loss on the ranking list $\mathbf{M}_{i}$, which can be formulated as:
\begin{equation}
\label{formula:cl_aux}
\begin{split}
\mathcal{L}_{\mathrm{RANK}} &= \frac{1}{|\mathbf{T}|} \sum_{k=1}^{|\mathbf{T}|} \sum_{\mathbf{h}_{i}^{(L)}, \mathbf{h}_{j}^{(L)} \in \mathbf{M}_{k}} \\ &\mathbb{I}_{\operatorname{r}(\mathbf{h}_{j}^{(L)}) > \operatorname{r}(\mathbf{h}_{i}^{(L)})} \log \left(1+\frac{s(\mathbf{q}^{\mathbf{s}}_{k},\mathbf{h}_{j}^{(L)})}{s(\mathbf{q}^{\mathbf{s}}_{k},\mathbf{h}_{i}^{(L)})}\right),
\end{split}
\end{equation}
where $\operatorname{r}(\cdot)$ denotes the ranking index (\textit{e.g.}, $\operatorname{r}(\mathbf{h}_{+}^{(L)}) < \operatorname{r}(\mathbf{h}_{1}^{(L)})$). 
Given $\mathbf{h}_{i}^{(L)}, \mathbf{h}_{j}^{(L)} \in \mathbf{M}_{k}$ that satisfies $\operatorname{r}(\mathbf{h}_{j}^{(L)}) > \operatorname{r}(\mathbf{h}_{i}^{(L)})$, the pair-wise ranking loss will explicitly optimize in the direction of ${\mathbf{E}_{\mathbf{q}}(\mathbf{q}^{\mathbf{s}}_{k})}^{\top} \mathbf{h}_{j}^{(L)} < {\mathbf{E}_{\mathbf{q}}(\mathbf{q}^{\mathbf{s}}_{k})}^{\top} \mathbf{h}_{i}^{(L)}$, thus imposing stricter constraints to the pair-wise ranking.

\textbf{Overall Training Objective}. 
To sum up, the overall training objective can be formulated as:
\begin{equation}
\mathcal{L} = \mathcal{L}_{\mathrm{CL}} + \alpha \cdot \mathcal{L}_{\mathrm{RANK}}.
\end{equation}
where $\alpha \in [0,1]$ is a hyper-parameter. 
It is worth noting that GoR's training costs are lightweight since the only trainable module is GNNs, and no human-crafted labels are needed.

\subsection{Retrieval from the Graph for Summarization}
\label{method:retrieval}

During the graph construction phase, we have already obtained a graph consisting of nodes that represent both the text chunks $\mathbf{c}$ from the long document and the responses $\mathbf{r}$ generated by LLMs during the RAG process.
These nodes collectively form the retrieval corpus used by GoR during inference.
After GNN training, each node is associated with a learned embedding vector that captures not only its semantic content but also its contextual and structural relationships with other nodes in the graph. These learned embeddings replace the original text embeddings produced by conventional retrievers (\textit{e.g.}, Contriever~\citep{izacard2021unsupervised-contriever}), enabling the model to incorporate richer relational information among document chunks, which is especially beneficial for the summarization task.
During inference for global summarization, the process begins with encoding the query into an embedding using the same retriever. We then compute the similarity (via inner product) between the query embedding and all node embeddings in the graph. The top-$\mathbf{K}$ most relevant nodes, comprising both document chunks and LLM-generated responses, are retrieved based on this similarity score. These selected nodes, along with the query, are then fed into the LLM to generate the final summary.

Overall, the retrieval process in GoR mirrors that of standard dense retrievers, with the key distinction being the use of graph-enhanced node embeddings and the inclusion of generated responses in the retrieval corpus. This approach allows GoR to better exploit both content and structural cues for improved summarization performance.

\section{Experiments}

\subsection{Experimental Setup}
\label{sec:setup}

\begin{table*}[t]
  \footnotesize
  \centering
  \begin{center}
  \begin{tabular}{ccccccccccccc}
    \toprule
    \multirow{2}{*}{Model} & \multicolumn{3}{c}{QMSum} & \multicolumn{3}{c}{AcademicEval} & \multicolumn{3}{c}{WCEP} & \multicolumn{3}{c}{BookSum} \\
    \cmidrule(r){2-4} \cmidrule(r){5-7} \cmidrule(r){8-10} \cmidrule(r){11-13}
    \multicolumn{1}{c}{} & R-L & R-1 & R-2 & R-L & R-1 & R-2 & R-L & R-1 & R-2 & R-L & R-1 & R-2 \\
    \midrule
    % Results
    Node2Vec
    & 18.5 & 31.8 & 6.3 
    & 19.3 & 38.3 & 10.6 
    & 13.9 & 20.1 & 6.3 
    & 13.6 & 27.4 & 4.6 \\
    \midrule
    BM25
    & 18.4 & 32.1 & 6.1 
    & 20.4 & 39.6 & 11.3 
    & 15.5 & 22.6 & 7.3 
    & 13.7 & 26.7 & 4.9 \\
    TF-IDF
    & 18.3 & 31.2 & 6.3 
    & 19.5 & 38.0 & 10.6 
    & 15.3 & 22.3 & 7.3 
    & 13.6 & 26.6 & 4.9 \\
    \midrule
    Contriever
    & 19.1 & 32.7 & 7.7 
    & 23.6 & 44.8 & 16.0 
    & 15.7 & 23.5 & 7.7 
    & 14.4 & 29.8 & 5.5 \\
    DPR
    & 18.6 & 32.1 & 6.7 
    & 20.9 & 41.4 & 13.2 
    & 15.6 & 22.5 & 7.5 
    & 13.8 & 27.1 & 4.8 \\
    Dragon
    & 19.2 & 33.5 & 7.7 
    & 23.5 & 43.8 & 15.1 
    & 14.6 & 21.8 & 6.8 
    & 13.7 & 27.2 & 4.8 \\
    SBERT
    & 19.0 & 33.0 & 7.4 
    & 23.4 & 45.2 & 15.8 
    & 13.7 & 20.5 & 5.5 
    & 14.4 & 29.5 & 5.4 \\
    \midrule
    BM25+DPR
    & 18.3 & 31.8 & 6.6 
    & 19.9 & 39.0 & 10.8 
    & 15.7 & 22.1 & 7.6 
    & 14.1 & 28.9 & 5.4 \\
    \midrule
    Gemma-8K
    & \textbf{19.8} & 33.5 & 7.3 
    & 21.9 & 42.0 & 12.9 
    & 15.6 & 21.9 & 7.7 
    & 12.8 & 23.4 & 4.2 \\
    Mistral-8K
    & 19.6 & 31.2 & 7.2 
    & 21.6 & 41.6 & 13.1 
    & 16.7 & 24.2 & 8.8 
    & 13.5 & 26.2 & 5.3 \\
    \midrule
    Full Context
    & 19.4 & 33.1 & 6.8 
    & 21.5 & 41.1 & 12.5 
    & 14.4 & 21.0 & 7.1 
    & 14.4 & 28.9 & 5.9 \\
    \midrule
    Thought-R
    & 19.0 & 33.9 & 7.6 
    & 22.0 & 42.6 & 13.2 
    & 15.2 & 22.4 & 7.4 
    & 14.2 & 29.5 & 5.7 \\
    \midrule
    \textbf{GoR (Ours)}
    & \textbf{19.8} & \textbf{34.5} & \textbf{7.8} 
    & \textbf{24.7} & \textbf{46.5} & \textbf{17.3} 
    & \textbf{18.1} & \textbf{25.4} & \textbf{9.2} 
    & \textbf{14.9} & \textbf{31.5} & \textbf{6.6} \\
    % Results
    \bottomrule
  \end{tabular}
  \end{center}
  \caption{\textbf{Experimental results on QMSum, AcademicEval, WCEP, and BookSum datasets over long-context global summarization tasks w.r.t. Rouge-L (R-L), Rouge-1 (R-1), and Rouge-2 (R-2)}. Note that the average LLM input token length of GoR and retriever-based baselines is $6 \times 256$ ($\approx$1.5K). (\textbf{BOLD} indicates the best score)}
    \label{tab:main_results}
\end{table*}

\textbf{Datasets}. 
We evaluate our proposed method on four long-context summarization datasets, \textit{i.e.}, AcademicEval~\citep{Thought-Retriever}, QMSum~\citep{QMSum}, WCEP~\citep{WCEP}, and BookSum~\citep{kryscinski2021booksum}. 
Among them, AcademicEval collects scientific papers from arXiv for abstract writing, given the long inputs of its main body. 
QMSum is a query-based summarization dataset, and we \textbf{only use} ``\textit{general queries}” for evaluating global summarization. 
WCEP is a multi-document summarization dataset about news events, while BookSum features long-form narrative summarization. 
For metrics, we adopt Rouge-1 (R-1), Rouge-2 (R-2), and Rouge-L  (R-L)~\citep{lin2004rouge} to assess the text alignment between the reference summaries and the predicted content generated by GoR.

\textbf{Implementation Details}. 
Following the standard procedure of RAG, we adopt TokenTextSplitter from LangChain to split each long document into text chunks. Each chunk has a size of 256, and the chunk overlapping is 32. 
We generate 30 queries for each long document using Mixtral-8x7B-Instruct-v0.1~\citep{jiang2024mixtral}, and the temperature coefficient is set to 0.5 by default in the query simulation stage. 
For RAG, we use Contriever~\citep{izacard2021unsupervised-contriever} for query and document embedding and retrieve 6 text chunks by default, which are fed into LLaMA-2-7b-chat~\citep{touvron2023llama-2} with greedy decoding to generate predicted summaries. In the training stage, we initialize the graph neural network as a two-layer graph attention network (GAT)~\citep{GAT}, with a 768-dim hidden dimension following the default setting of most retrievers.

\textbf{Baselines}. 
To have a comprehensive evaluation, we compare our proposed GoR with dozens of baselines, including (1) \textbf{Random Walk based Node Embedding} (\textit{i.e.}, Node2Vec~\citep{grover2016node2vec}), (2) \textbf{Sparse Retriever} (\textit{i.e.}, BM25~\citep{robertson2009probabilistic-bm25} and TF-IDF~\citep{ramos2003using-tf-idf}), (3) \textbf{Dense Retriever} (\textit{i.e.}, Contriever~\citep{izacard2021unsupervised-contriever}, DPR~\citep{karpukhin2020dense-dpr}, Dragon~\citep{lin2023train-dragon}, and Sentence-BERT (SBERT)~\citep{Sentence-BERT}\footnote{\space We use all-MiniLM-L6-v2 as the backbone.}), (4) \textbf{Hybrid Retriever} (\textit{i.e.}, BM25+DPR with Reciprocal Rerank Fusion), (5) \textbf{Long-context LLMs} (\textit{i.e.}, Gemma-8K~\citep{team2024gemma} and Mistral-8K~\citep{jiang2023mistral}), (6) \textbf{Full Context} (\textit{i.e.}, feeds all inputs to LLMs for summary generation\footnote{\space If the input length exceeds the context window limit, we randomly sample continuous text spans of maximum length multiple times to feed into LLMs and calculate the avg. result.}), and (7) \textbf{Thought Retriever} (Thought-R)~\citep{Thought-Retriever}. 
Appendix~\ref{appendix:exp_details} elucidates more details. 
% More experimental details can be found in Appendix~\ref{appendix:exp_details}. 

\subsection{Main Results}

\begin{table*}[ht]
\footnotesize
\centering
\begin{tabular}{cccccccc}
\toprule
\multicolumn{2}{c}{QMSum} & \multicolumn{2}{c}{AcademicEval} & \multicolumn{2}{c}{WCEP} & \multicolumn{2}{c}{BookSum} \\ 
\cmidrule(r){1-2} \cmidrule(r){3-4} \cmidrule(r){5-6} \cmidrule(r){7-8} BM25 & \textbf{GoR} & BM25 & \textbf{GoR} & \textbf{BM25} & GoR & BM25 & \textbf{GoR} \\
38.7\%      & \textbf{61.3\%}     & 30.0\%      & \textbf{70.0\%}  & \textbf{70.4\%}      & 29.6\% & 35.5\%      & \textbf{64.5\%} \\
\cmidrule(r){1-2} \cmidrule(r){3-4} \cmidrule(r){5-6} \cmidrule(r){7-8} TF-IDF & \textbf{GoR} & TF-IDF & \textbf{GoR} & \textbf{TF-IDF} & GoR & TF-IDF & \textbf{GoR} \\
40.6\%      & \textbf{59.4\%}     & 33.3\%      & \textbf{66.7\%}  & \textbf{66.7\%}      & 33.3\% & 41.9\%      & \textbf{58.1\%} \\
\cmidrule(r){1-2} \cmidrule(r){3-4} \cmidrule(r){5-6} \cmidrule(r){7-8} \textbf{Contriever} & GoR & \textbf{Contriever} & GoR & Contriever & \textbf{GoR} & \textbf{Contriever} & GoR \\
\textbf{51.6\%}      & 48.4\%    & \textbf{53.3\%}      & 46.7\% & 48.1\%      & \textbf{51.9\%} & \textbf{54.8\%}      & 45.2\% \\
\cmidrule(r){1-2} \cmidrule(r){3-4} \cmidrule(r){5-6} \cmidrule(r){7-8} Gemma-8K & \textbf{GoR} & Gemma-8K & \textbf{GoR} & Gemma-8K & \textbf{GoR} & Gemma-8K & \textbf{GoR} \\
6.5\%      & \textbf{93.5\%}     & 16.7\%      & \textbf{83.3\%} & 37.0\%      & \textbf{63.0\%} & 12.9\%      & \textbf{87.1\%} \\
\cmidrule(r){1-2} \cmidrule(r){3-4} \cmidrule(r){5-6} \cmidrule(r){7-8} Mistral-8K & \textbf{GoR} & Mistral-8K & \textbf{GoR} & Mistral-8K & \textbf{GoR} & Mistral-8K & \textbf{GoR} \\
12.9\%      & \textbf{87.1\%}     & 40.0\%      & \textbf{60.0\%} & 37.0\%      & \textbf{63.0\%} & 22.6\%      & \textbf{77.4\%} \\
\bottomrule
\end{tabular}
\caption{\textbf{LLM Evaluation w.r.t. overall win rates on the QMSum, AcademicEval, WCEP, and BookSum datasets.} Note that there are very few test samples that contain some security-sensitive information that causes DeepSeek-R1 to be unable to return valid evaluation information. We directly skip these samples.}
\label{tab:llm_eval}
\end{table*}

We conduct comprehensive experiments on QMSum, AcademicEval, WCEP, and BookSum datasets compared with dozens of baselines to evaluate the long-context global summarization capabilities of our proposed method. The results are shown in Table~\ref{tab:main_results}. 

\textbf{GoR consistently outperforms retriever-based methods}. 
From Table~\ref{tab:main_results}, our proposed GoR beats sparse retrievers, dense retrievers, and hybrid retrievers in every aspect. 
Thanks to the constructed graph, which integrates text chunks from long documents and LLM historical responses into a whole, node embeddings can better reflect the complicated correlations with given queries, thus significantly improving the retrieval performance of GoR. 
Moreover, the informative content of historical responses may also enhance the summarization task. 

\textbf{GoR shows superiority over long-context LLMs}. 
We compare Gemma-8K and Mistral-8K with a longer context window to accommodate long-context inputs. However, longer inputs may contain minor information, and long-context LLMs struggle with this situation. In contrast, GoR can effectively differentiate key and topic-related content in long texts using learned node embeddings and achieve better results with shorter input lengths. 

\textbf{Additional Findings}. 
(1) Node2Vec produces unsatisfactory results, and the node embeddings cannot be optimized effectively since it is based on a non-differentiable algorithm. 
(2) Although Thought Retriever demonstrates competitive results, it is still inferior to GoR due to the lack of exploration of the correlations between retrieved text chunks and LLM-generated responses. 
(3) Since the context window length limit of LLMs is exceeded, “Full Context” truncates the long-context input, thus losing some information that may be important for global summarization, resulting in suboptimal results. 

Overall, GoR achieves the best results compared with various baselines, demonstrating the effectiveness of our proposed method.

\subsection{LLM Evaluation}
\label{exp:llm_eval}

To enable a more comprehensive automatic evaluation of GoR, inspired by LLM-as-a-Judge~\citep{gu2024surveyllmasajudge}, we adopt DeepSeek-R1~\citep{guo2025deepseekr1} to assess the summaries generated by GoR and competitive baselines. 
Following~\citep{GraphRAG} and~\citep{guo2024lightrag}, we evaluate from three perspectives: comprehensiveness, diversity, and empowerment. The LLM is instructed to provide an overall judgment based on these criteria to determine the better summary.

To reduce evaluation costs, we select representative and strong baselines for LLM evaluation on the QMSum, AcademicEval, WCEP, and BookSum datasets.
Table~\ref{tab:llm_eval} reports the overall win rates, \textit{i.e.}, the proportion of summaries judged better under pairwise comparison. Evaluation prompts are provided in Appendix~\ref{appendix:prompts}.

From Table~\ref{tab:llm_eval}, we observe the following. 
(1) GoR consistently outperforms other methods across all datasets, demonstrating stronger comprehensiveness, diversity, and informativeness.
(2) Contriever performs comparably, with results close to GoR.
(3) Long-context LLMs like Gemma-8K and Mistral-8K fall significantly behind, suggesting that GoR’s graph-based retrieval yields more relevant and refined content, especially when input length is limited.

In summary, GoR delivers superior performance in LLM evaluation compared to other baselines.

\subsection{Ablation Study}

To investigate how each component of GoR contributes to its performance, we conduct an ablation experiment, and the results are shown in Table~\ref{tab:ablation}. 

From Table~\ref{tab:ablation}, we can draw several conclusions. 
(1) Directly using the text embeddings from the retriever without training leads to degraded performance (\textit{i.e.}, w/o train), highlighting the effectiveness of the learned node embeddings. 
(2) Both the contrastive loss $\mathcal{L}_{\mathrm{CL}}$ and pair-wise ranking loss $\mathcal{L}_{\mathrm{RANK}}$ significantly improve performance. The pair-wise ranking loss imposes stricter ranking constraints on node embeddings, making effective use of the indirect supervision signal from the self-supervised reference summaries. 
(3) In-batch negatives are crucial to the performance of contrastive learning. Removing in-batch negatives (\textit{i.e.}, w/o in-b neg) leads to a significant drop in results. 
(4) Compared with self-supervised training, we utilize global reference summaries as labels to conduct supervised training (\textit{i.e.}, w/ sup), and the results are significantly worse than the self-supervised setting. We will further discuss it in Section~\ref{sec:exp_sup}. 

In general, GoR's reasonable module design enables it to achieve superior performance. 

\begin{table}[ht]
  \footnotesize
    \centering
  \begin{center}
  \begin{tabular}{ccccccc}
    \toprule
    \multirow{2}{*}{Variant} & \multicolumn{3}{c}{WCEP} & \multicolumn{3}{c}{BookSum} \\
    \cmidrule(r){2-4} \cmidrule(r){5-7}
    \multicolumn{1}{c}{} & R-L & R-1 & R-2 & R-L & R-1 & R-2 \\
    \midrule
    % Results
    w/o train
    & 15.3 & 22.4 & 7.4 
    & 13.7 & 27.7 & 4.7 \\
    w/o $\mathcal{L}_{\mathrm{CL}}$
    & 14.7 & 21.9 & 7.2 
    & 14.1 & 28.8 & 5.1 \\
    w/o $\mathcal{L}_{\mathrm{RANK}}$
    & 16.6 & 24.2 & 8.2 
    & 14.0 & 28.0 & 4.9 \\
    w/o in-b neg
    & 17.2 & 24.9 & 8.8 
    & 13.3 & 26.3 & 5.2 \\
    w/ sup
    & 15.5 & 22.8 & 7.3 
    & 13.8 & 29.0 & 5.2 \\
    \midrule
    \textbf{GoR}
    & \textbf{18.1} & \textbf{25.4} & \textbf{9.2} 
    & \textbf{14.9} & \textbf{31.5} & \textbf{6.6} \\
    % Results
    \bottomrule
  \end{tabular}
  \end{center}
  \caption{\textbf{Ablation study on the WCEP and BookSum datasets w.r.t. R-L, R-1, and R-2}.}
    \label{tab:ablation}
\end{table}

\subsection{Discussions}
\label{sec:exp_sup}

\begin{figure}[h]
\begin{center}
    \centering
    \begin{subfigure}{0.49\linewidth}
        \centering
        \includegraphics[width=1.0\linewidth]{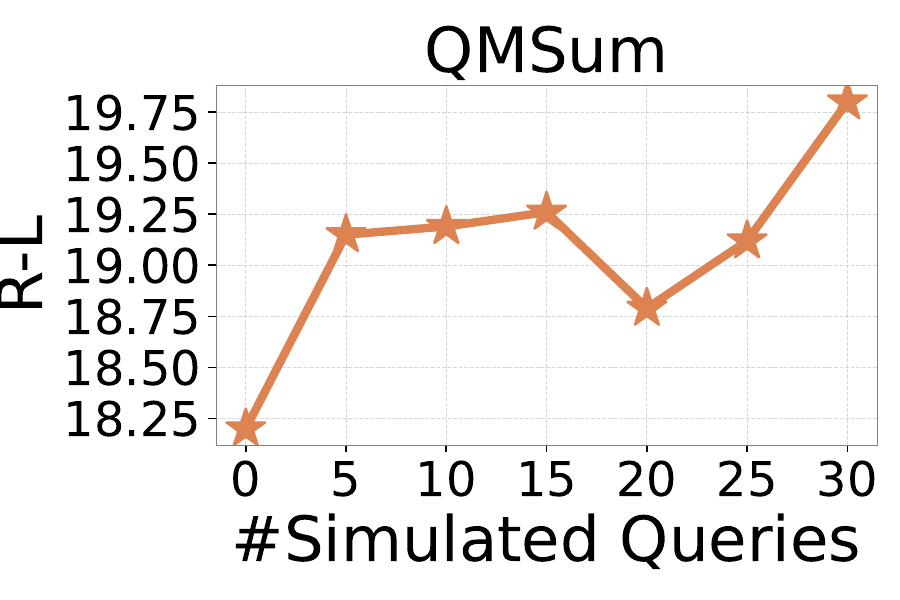}
        \label{fig:query_qmsum}
    \end{subfigure}
    \centering
    \begin{subfigure}{0.49\linewidth}
        \centering
        \includegraphics[width=1.0\linewidth]{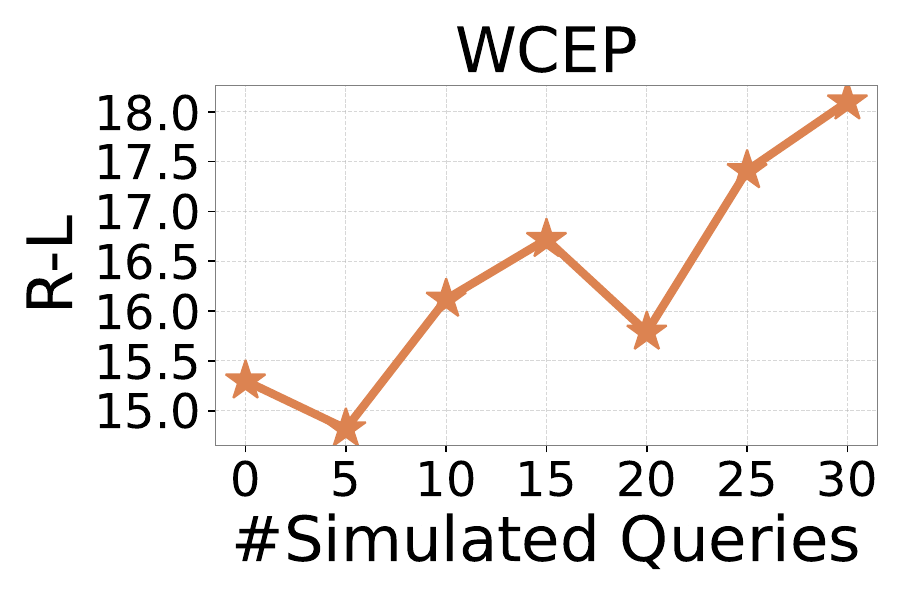}
        \label{fig:query_wcep}
    \end{subfigure}
    \caption{\textbf{Impact of the number of simulated queries during training w.r.t. R-L}. We show the results on the QMSum and WCEP datasets.}
    \label{fig:query}
\end{center}
\end{figure}

\textbf{Impact of the Number of Simulated Queries During Training}. 
Query Simulation is a crucial stage in our method design, and we will examine how the number of simulated queries used during training affects learning performance. In particular, we explore this effect by gradually increasing the number of simulated queries used in training. 
We present the results in Figure~\ref{fig:query}. Overall, R-L shows an upward trend as the number of simulated queries increases. Nevertheless, since fewer queries cover less relevant content from long documents, the curves of each dataset have some fluctuations, indicating the occurrence of underfitting. 

In general, 30 simulated queries can optimize the model well across these four datasets, which indicates that our proposed GoR is cost-effective. 
Nevertheless, increasing the number of simulated queries may still potentially further improve the performance of the model. Due to budget constraints, we will leave this for future work.

% Hyper Table
\begin{table*}[t]
  \footnotesize
  \centering
%   \vspace{-1ex}
\begin{center}
  \begin{tabular}{cccccccc}
    \toprule
    Baselines & \makecell{Node2Vec} & \makecell{BM25} & \makecell{Contriever} & \makecell{SBERT} & \makecell{BM25+DPR} & \makecell{Thought-R} & \makecell{\textbf{GoR (ours)}} \\
    \midrule
    % Results
    \makecell{Inference\\Time (s)}
    & 9.4 & 0.02 & 0.20 & 0.01 & 0.04 & 0.3 & 0.58 \\
    % Results
    \bottomrule
  \end{tabular}
\end{center}
\caption{\textbf{Inference efficiency analysis w.r.t. inference time per query on the WCEP dataset.}}
  \label{tab:inference}
\end{table*}

\begin{figure}[h]
\begin{center}
    \centering
    \begin{minipage}{0.49\linewidth}
        \includegraphics[width=\linewidth]{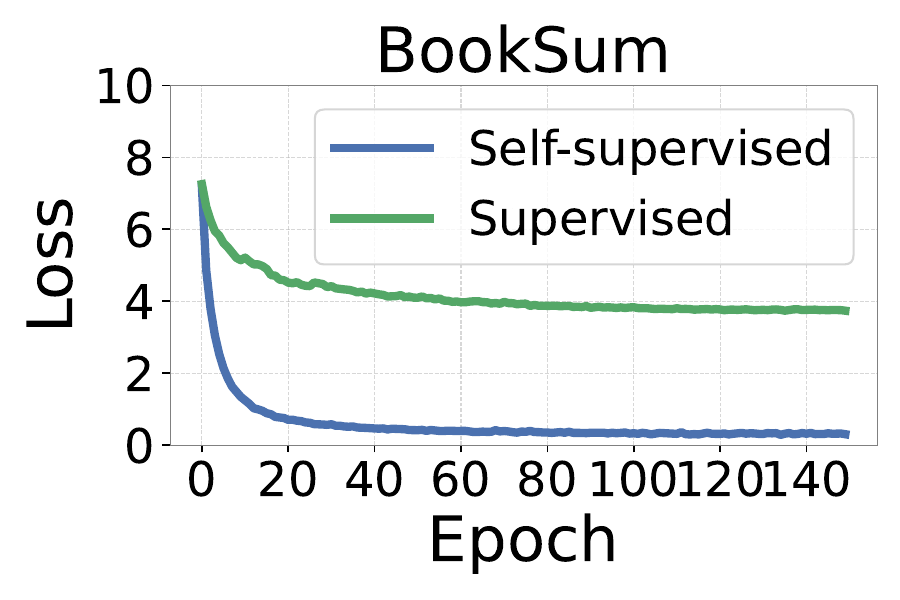}
        \label{fig:loss_book}
    \end{minipage}
    \hfill
    \begin{minipage}{0.49\linewidth}
        \includegraphics[width=\linewidth]{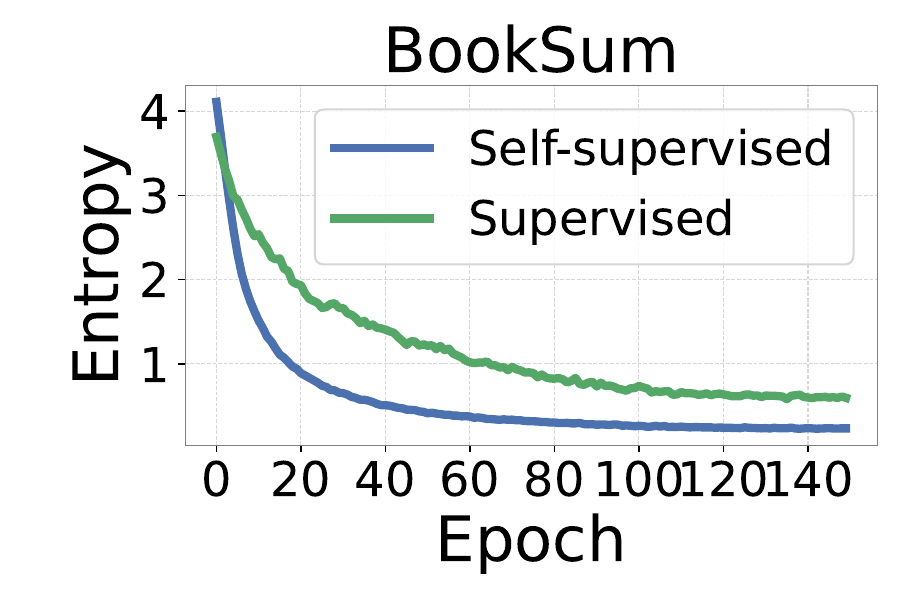}
        \label{fig:entropy_book}
    \end{minipage}
    \caption{\textbf{Differences between self-supervised and supervised training w.r.t. loss and entropy on the BookSum dataset.}}
    \label{fig:loss_entropy}
\end{center}
\end{figure}

\textbf{Supervised Training on Global Summarization Queries}. 
To dive deeper into the differences between self-supervised and supervised training, we carry out additional experiments using global reference summaries. 
Specifically, we utilize global summarization queries and reference summaries to serve as a training corpus under the supervised setting. 
As there is only one global summarization query for each long document, we replicate it multiple times to match the quantity of self-supervised training data, thus eliminating the impact of the quantity difference. 
We present the results on the BookSum dataset in Figure~\ref{fig:loss_entropy}, and the \textit{Entropy} denotes the entropy of the similarity distribution between queries and node embeddings. 

From Figure~\ref{fig:loss_entropy}, it is evident that in the self-supervised setting, the loss is consistently lower than in the supervised setting. 
This suggests that the global reference summaries are highly correlated with many nodes, causing most nodes to exhibit a high semantic similarity with the global query. 
As a result, this confuses the model's optimization direction. 
Additionally, the entropy curve shows that the entropy in the supervised setting is consistently higher than in the self-supervised setting, indicating that the model struggles to select the most similar node. 
In contrast, the self-supervised label, derived from a specific part of a long document, contains more focused content, effectively guiding the model's optimization.

\textbf{Inference Efficiency Analysis}. 
To investigate the inference efficiency of GoR, we conduct extensive experiments on the WCEP dataset and present the results w.r.t. the inference time per query in Table~\ref{tab:inference}. Note that since the LLM used in our experiment is consistent, we ignore the inference time brought by the LLM itself.

From Table~\ref{tab:inference}, we can draw the following conclusion. 
(1) Since GoR's only trainable module is GNN, GoR's inference efficiency is very high, and almost no additional noticeable latency is introduced. 
(2) Although GoR's inference time is longer than some baselines, it only increases by a few hundred milliseconds. Considering the significant performance improvement brought by GoR in Table~\ref{tab:main_results}, this tiny time overhead is almost negligible in practical applications.

\section{Related Work}

\textbf{Long-context Summarization using LLMs.} 
In recent years, LLMs have shown impressive capabilities in long-context modeling~\citep{achiam2023GPT-4, llama3modelcard, team2024gemma, jiang2024mixtral}. 
Summarizing lengthy documents with LLMs primarily involves two approaches: retrieval-augmented generation (RAG)~\citep{RALM, yu2023ralm} and long-context LLMs~\citep{glm2024chatglm, longchat2023, longllama}. 
Long-context LLMs feature a large context window length to accommodate long-context inputs. 
However, they may suffer from severe performance degradation when accessing some local key details in the middle of long contexts~\citep{liu2024lost-middle}. 
Conversely, RAG emerges as a promising approach for cost-effective long-context summarization. 
By equipping with a retriever~\citep{karpukhin2020dense-dpr, robertson2009probabilistic-bm25}, RAG can first perform a relevance search based on user queries and then feed the retrieved text into LLMs for summary. 
For a recent example, GraphRAG~\citep{GraphRAG} conducts query-focused summarization by setting up a graph index and detecting graph communities for summary generation. 

Nevertheless, most current RAG approaches still focus on enhancing LLMs' reasoning and question-answering capabilities, which only require retrieving locally relevant information~\citep{trivedi2022interleaving-ralm, jiang2023active-ralm, asai2023self-rag, CoK, Takeastepback}. 
In comparison, our proposed method stands out from these methods by focusing on LLMs' global summarization capability of long-context inputs.

\textbf{Graph-assisted Retrieval-augmented Language Models.} 
As one of the effective structures for modeling data relations, graphs have recently been used to enhance the performance of retrieval-augmented language models on various QA tasks. 
EtD~\citep{EtD} features a graph neural network (GNN) to traverse a knowledge graph hop by hop to discover more relevant knowledge, thus enhancing LLM generation quality. 
GNN-RAG~\citep{GNN-RAG} learns to reason over graphs using GNNs, and retrieves answer candidates for a given question.
PG-RAG~\citep{PG-RAG} constructs pseudo-graphs with a retrieval indexer by prompting LLMs to organize document knowledge in a self-learning manner. 
G-RAG~\citep{G-RAG} proposes to rerank documents by learning graph representation on abstract meaning representation graphs, while GNN-Ret~\citep{GNN-Ret} refines semantic distances between documents and queries by modeling relationships among related passages. 
ToG~\citep{ToG} and KGP~\citep{KGP} treat LLMs as agents to traverse and reason over knowledge graphs in an iterative way, while RoG~\citep{RoG} first generates plans for retrieval and then conducts reasoning. 
G-Retriever~\citep{G-retriever} transforms retrieved knowledge subgraphs into graph embeddings by training a graph encoder and textualizes subgraphs to serve as inputs of LLMs. 

Different from the above, GraphRAG~\citep{GraphRAG} sets up a graph index for query-focused summarization, which aligns more closely with our approach. 
Compared with GraphRAG, which is time-consuming and suffers from huge computational costs, GoR is lightweight and only draws on a few LLM historical responses with efficient training to achieve competitive performance.

\section{Conclusion}

In this work, we introduce a method named \textit{graph of records} (GoR) to improve long-context global summarization in retrieval-augmented generation by utilizing LLM-generated historical responses. 
Intuitively, we establish connections between text chunks retrieved from long documents and LLM-generated historical responses to create a graph of records. 
To uncover complex correlations between these connections, we use a graph neural network and develop a BERTScore-based objective for self-supervised training, enabling seamless supervision signal backpropagation between self-supervised reference summaries and node embeddings. 
Our experiments on four long-context summarization datasets show that GoR significantly outperforms various baselines, demonstrating its effectiveness.

\section{Limitations}

Despite the superiority of our proposed method, GoR has some limitations. 
(1) Due to a limited budget, we only simulate and generate a small number of user queries, which may cause a bottleneck in further model optimization. 
(2) The simulated queries may not accurately reflect the real-world distribution, as they do not account for the possibility of users asking many meaningless questions. Therefore, a filtering process may be necessary, which we leave for future work.

To promote sharing and communication in the academic community, we also share some insights about simulated queries here. 
Given the powerful evaluation capabilities of LLMs, many works utilize LLM-as-a-Judge and treat LLMs as evaluators. Intuitively, in practical applications, we can first use some simple rule-based filtering strategies to preliminarily screen the meaningless questions raised by users, and then use LLM-as-a-Judge to evaluate the remaining questions and judge their quality from multiple dimensions (such as diversity, complexity, inspiration, etc.), which not only takes into account performance but also ensures a certain efficiency in real multi-user scenarios.

\section*{Acknowledgments}

We sincerely appreciate the support from Amazon grant funding project \#120359, "GRAG: Enhance RAG Applications with Graph-structured Knowledge", and Meta gift funding project "PERM: Toward Parameter Efficient Foundation Models for Recommenders"

% Bibliography entries for the entire Anthology, followed by custom entries
%\bibliography{anthology,custom}
% Custom bibliography entries only
\bibliography{anthology,custom}

\begin{thebibliography}{62}
\providecommand{\natexlab}[1]{#1}

\bibitem[{Achiam et~al.(2023)Achiam, Adler, Agarwal, Ahmad, Akkaya, Aleman, Almeida, Altenschmidt, Altman, Anadkat et~al.}]{achiam2023GPT-4}
Josh Achiam, Steven Adler, Sandhini Agarwal, Lama Ahmad, Ilge Akkaya, Florencia~Leoni Aleman, Diogo Almeida, Janko Altenschmidt, Sam Altman, Shyamal Anadkat, et~al. 2023.
\newblock Gpt-4 technical report.
\newblock \emph{arXiv preprint arXiv:2303.08774}.

\bibitem[{AI@Meta(2024)}]{llama3modelcard}
AI@Meta. 2024.
\newblock \href {https://github.com/meta-llama/llama3/blob/main/MODEL_CARD.md} {Llama 3 model card}.

\bibitem[{Asai et~al.(2023)Asai, Wu, Wang, Sil, and Hajishirzi}]{asai2023self-rag}
Akari Asai, Zeqiu Wu, Yizhong Wang, Avirup Sil, and Hannaneh Hajishirzi. 2023.
\newblock Self-rag: Learning to retrieve, generate, and critique through self-reflection.
\newblock \emph{arXiv preprint arXiv:2310.11511}.

\bibitem[{Besta et~al.(2024)Besta, Blach, Kubicek, Gerstenberger, Podstawski, Gianinazzi, Gajda, Lehmann, Niewiadomski, Nyczyk et~al.}]{GoT}
Maciej Besta, Nils Blach, Ales Kubicek, Robert Gerstenberger, Michal Podstawski, Lukas Gianinazzi, Joanna Gajda, Tomasz Lehmann, Hubert Niewiadomski, Piotr Nyczyk, et~al. 2024.
\newblock Graph of thoughts: Solving elaborate problems with large language models.
\newblock In \emph{Proceedings of the AAAI Conference on Artificial Intelligence}, volume~38, pages 17682--17690.

\bibitem[{Burges(2010)}]{lambdarank}
Christopher~JC Burges. 2010.
\newblock From ranknet to lambdarank to lambdamart: An overview.
\newblock \emph{Learning}, 11(23-581):81.

\bibitem[{Dong et~al.(2024)Dong, Fatemi, Perozzi, Yang, and Tsitsulin}]{G-RAG}
Jialin Dong, Bahare Fatemi, Bryan Perozzi, Lin~F Yang, and Anton Tsitsulin. 2024.
\newblock Don't forget to connect! improving rag with graph-based reranking.
\newblock \emph{arXiv preprint arXiv:2405.18414}.

\bibitem[{Edge et~al.(2024)Edge, Trinh, Cheng, Bradley, Chao, Mody, Truitt, and Larson}]{GraphRAG}
Darren Edge, Ha~Trinh, Newman Cheng, Joshua Bradley, Alex Chao, Apurva Mody, Steven Truitt, and Jonathan Larson. 2024.
\newblock From local to global: A graph rag approach to query-focused summarization.
\newblock \emph{arXiv preprint arXiv:2404.16130}.

\bibitem[{Feng et~al.(2024)Feng, Han, Lin, Liu, and You}]{Thought-Retriever}
Tao Feng, Pengrui Han, Guanyu Lin, Ge~Liu, and Jiaxuan You. 2024.
\newblock Thought-retriever: Don’t just retrieve raw data, retrieve thoughts.
\newblock In \emph{International Conference on Learning Representations Workshop: How Far Are We From AGI}.

\bibitem[{Gholipour~Ghalandari et~al.(2020)Gholipour~Ghalandari, Hokamp, Pham, Glover, and Ifrim}]{WCEP}
Demian Gholipour~Ghalandari, Chris Hokamp, Nghia~The Pham, John Glover, and Georgiana Ifrim. 2020.
\newblock \href {https://www.aclweb.org/anthology/2020.acl-main.120} {A large-scale multi-document summarization dataset from the {W}ikipedia current events portal}.
\newblock In \emph{Proceedings of the 58th Annual Meeting of the Association for Computational Linguistics}, pages 1302--1308, Online. Association for Computational Linguistics.

\bibitem[{GLM et~al.(2024)GLM, Zeng, Xu, Wang, Zhang, Yin, Rojas, Feng, Zhao, Lai, Yu, Wang, Sun, Zhang, Cheng, Gui, Tang, Zhang, Li, Zhao, Wu, Zhong, Liu, Huang, Zhang, Zheng, Lu, Duan, Zhang, Cao, Yang, Tam, Zhao, Liu, Xia, Zhang, Gu, Lv, Liu, Liu, Yang, Song, Zhang, An, Xu, Niu, Yang, Li, Bai, Dong, Qi, Wang, Yang, Du, Hou, and Wang}]{glm2024chatglm}
Team GLM, Aohan Zeng, Bin Xu, Bowen Wang, Chenhui Zhang, Da~Yin, Diego Rojas, Guanyu Feng, Hanlin Zhao, Hanyu Lai, Hao Yu, Hongning Wang, Jiadai Sun, Jiajie Zhang, Jiale Cheng, Jiayi Gui, Jie Tang, Jing Zhang, Juanzi Li, Lei Zhao, Lindong Wu, Lucen Zhong, Mingdao Liu, Minlie Huang, Peng Zhang, Qinkai Zheng, Rui Lu, Shuaiqi Duan, Shudan Zhang, Shulin Cao, Shuxun Yang, Weng~Lam Tam, Wenyi Zhao, Xiao Liu, Xiao Xia, Xiaohan Zhang, Xiaotao Gu, Xin Lv, Xinghan Liu, Xinyi Liu, Xinyue Yang, Xixuan Song, Xunkai Zhang, Yifan An, Yifan Xu, Yilin Niu, Yuantao Yang, Yueyan Li, Yushi Bai, Yuxiao Dong, Zehan Qi, Zhaoyu Wang, Zhen Yang, Zhengxiao Du, Zhenyu Hou, and Zihan Wang. 2024.
\newblock \href {https://arxiv.org/abs/2406.12793} {Chatglm: A family of large language models from glm-130b to glm-4 all tools}.
\newblock \emph{Preprint}, arXiv:2406.12793.

\bibitem[{Grover and Leskovec(2016)}]{grover2016node2vec}
Aditya Grover and Jure Leskovec. 2016.
\newblock node2vec: Scalable feature learning for networks.
\newblock In \emph{Proceedings of the 22nd ACM international conference on Knowledge discovery and data mining}, pages 855--864.

\bibitem[{Gu et~al.(2024)Gu, Jiang, Shi, Tan, Zhai, Xu, Li, Shen, Ma, Liu, Wang, and Guo}]{gu2024surveyllmasajudge}
Jiawei Gu, Xuhui Jiang, Zhichao Shi, Hexiang Tan, Xuehao Zhai, Chengjin Xu, Wei Li, Yinghan Shen, Shengjie Ma, Honghao Liu, Yuanzhuo Wang, and Jian Guo. 2024.
\newblock A survey on llm-as-a-judge.
\newblock \emph{arXiv preprint arXiv: 2411.15594}.

\bibitem[{Guo et~al.(2025)Guo, Yang, Zhang, Song, Zhang, Xu, Zhu, Ma, Wang, Bi et~al.}]{guo2025deepseekr1}
Daya Guo, Dejian Yang, Haowei Zhang, Junxiao Song, Ruoyu Zhang, Runxin Xu, Qihao Zhu, Shirong Ma, Peiyi Wang, Xiao Bi, et~al. 2025.
\newblock Deepseek-r1: Incentivizing reasoning capability in llms via reinforcement learning.
\newblock \emph{arXiv preprint arXiv:2501.12948}.

\bibitem[{Guo et~al.(2024)Guo, Xia, Yu, Ao, and Huang}]{guo2024lightrag}
Zirui Guo, Lianghao Xia, Yanhua Yu, Tu~Ao, and Chao Huang. 2024.
\newblock Lightrag: Simple and fast retrieval-augmented generation.
\newblock \emph{arXiv preprint arXiv:2410.05779}.

\bibitem[{Hamilton et~al.(2017)Hamilton, Ying, and Leskovec}]{GraphSAGE}
Will Hamilton, Zhitao Ying, and Jure Leskovec. 2017.
\newblock Inductive representation learning on large graphs.
\newblock In \emph{Annual Conference on Neural Information Processing Systems}.

\bibitem[{He et~al.(2024)He, Tian, Sun, Chawla, Laurent, LeCun, Bresson, and Hooi}]{G-retriever}
Xiaoxin He, Yijun Tian, Yifei Sun, Nitesh~V Chawla, Thomas Laurent, Yann LeCun, Xavier Bresson, and Bryan Hooi. 2024.
\newblock G-retriever: Retrieval-augmented generation for textual graph understanding and question answering.
\newblock \emph{arXiv preprint arXiv:2402.07630}.

\bibitem[{Huang et~al.(2021)Huang, Cao, Parulian, Ji, and Wang}]{GovReport}
Luyang Huang, Shuyang Cao, Nikolaus Parulian, Heng Ji, and Lu~Wang. 2021.
\newblock \href {https://arxiv.org/abs/2104.02112} {Efficient attentions for long document summarization}.
\newblock \emph{Preprint}, arXiv:2104.02112.

\bibitem[{Izacard et~al.(2021)Izacard, Caron, Hosseini, Riedel, Bojanowski, Joulin, and Grave}]{izacard2021unsupervised-contriever}
Gautier Izacard, Mathilde Caron, Lucas Hosseini, Sebastian Riedel, Piotr Bojanowski, Armand Joulin, and Edouard Grave. 2021.
\newblock Unsupervised dense information retrieval with contrastive learning.
\newblock \emph{arXiv preprint arXiv:2112.09118}.

\bibitem[{Jiang et~al.(2023{\natexlab{a}})Jiang, Sablayrolles, Mensch, Bamford, Chaplot, Casas, Bressand, Lengyel, Lample, Saulnier et~al.}]{jiang2023mistral}
Albert~Q Jiang, Alexandre Sablayrolles, Arthur Mensch, Chris Bamford, Devendra~Singh Chaplot, Diego de~las Casas, Florian Bressand, Gianna Lengyel, Guillaume Lample, Lucile Saulnier, et~al. 2023{\natexlab{a}}.
\newblock Mistral 7b.
\newblock \emph{arXiv preprint arXiv:2310.06825}.

\bibitem[{Jiang et~al.(2024)Jiang, Sablayrolles, Roux, Mensch, Savary, Bamford, Chaplot, Casas, Hanna, Bressand et~al.}]{jiang2024mixtral}
Albert~Q Jiang, Alexandre Sablayrolles, Antoine Roux, Arthur Mensch, Blanche Savary, Chris Bamford, Devendra~Singh Chaplot, Diego de~las Casas, Emma~Bou Hanna, Florian Bressand, et~al. 2024.
\newblock Mixtral of experts.
\newblock \emph{arXiv preprint arXiv:2401.04088}.

\bibitem[{Jiang et~al.(2023{\natexlab{b}})Jiang, Xu, Gao, Sun, Liu, Dwivedi-Yu, Yang, Callan, and Neubig}]{jiang2023active-ralm}
Zhengbao Jiang, Frank~F Xu, Luyu Gao, Zhiqing Sun, Qian Liu, Jane Dwivedi-Yu, Yiming Yang, Jamie Callan, and Graham Neubig. 2023{\natexlab{b}}.
\newblock Active retrieval augmented generation.
\newblock \emph{arXiv preprint arXiv:2305.06983}.

\bibitem[{Karpukhin et~al.(2020)Karpukhin, O{\u{g}}uz, Min, Lewis, Wu, Edunov, Chen, and Yih}]{karpukhin2020dense-dpr}
Vladimir Karpukhin, Barlas O{\u{g}}uz, Sewon Min, Patrick Lewis, Ledell Wu, Sergey Edunov, Danqi Chen, and Wen-tau Yih. 2020.
\newblock Dense passage retrieval for open-domain question answering.
\newblock \emph{arXiv preprint arXiv:2004.04906}.

\bibitem[{Kipf and Welling(2016)}]{GCN}
Thomas~N Kipf and Max Welling. 2016.
\newblock Semi-supervised classification with graph convolutional networks.
\newblock \emph{arXiv preprint arXiv:1609.02907}.

\bibitem[{Kojima et~al.(2022)Kojima, Gu, Reid, Matsuo, and Iwasawa}]{kojima2022zero-shot-cot}
Takeshi Kojima, Shixiang~Shane Gu, Machel Reid, Yutaka Matsuo, and Yusuke Iwasawa. 2022.
\newblock Large language models are zero-shot reasoners.
\newblock \emph{Advances in neural information processing systems}, 35:22199--22213.

\bibitem[{Kry{\'s}ci{\'n}ski et~al.(2021)Kry{\'s}ci{\'n}ski, Rajani, Agarwal, Xiong, and Radev}]{kryscinski2021booksum}
Wojciech Kry{\'s}ci{\'n}ski, Nazneen Rajani, Divyansh Agarwal, Caiming Xiong, and Dragomir Radev. 2021.
\newblock Booksum: A collection of datasets for long-form narrative summarization.
\newblock \emph{arXiv preprint arXiv:2105.08209}.

\bibitem[{Li* et~al.(2023)Li*, Shao*, Xie, Sheng, Zheng, Gonzalez, Stoica, Ma, and Zhang}]{longchat2023}
Dacheng Li*, Rulin Shao*, Anze Xie, Ying Sheng, Lianmin Zheng, Joseph~E. Gonzalez, Ion Stoica, Xuezhe Ma, and Hao Zhang. 2023.
\newblock \href {https://lmsys.org/blog/2023-06-29-longchat} {How long can open-source llms truly promise on context length?}

\bibitem[{Li et~al.(2024{\natexlab{a}})Li, Zhang, Do, Yue, and Chen}]{li2024long-context-struggle}
Tianle Li, Ge~Zhang, Quy~Duc Do, Xiang Yue, and Wenhu Chen. 2024{\natexlab{a}}.
\newblock Long-context llms struggle with long in-context learning.
\newblock \emph{arXiv preprint arXiv:2404.02060}.

\bibitem[{Li et~al.(2023)Li, Zhao, Chia, Ding, Joty, Poria, and Bing}]{CoK}
Xingxuan Li, Ruochen Zhao, Yew~Ken Chia, Bosheng Ding, Shafiq Joty, Soujanya Poria, and Lidong Bing. 2023.
\newblock Chain-of-knowledge: Grounding large language models via dynamic knowledge adapting over heterogeneous sources.
\newblock \emph{arXiv preprint arXiv:2305.13269}.

\bibitem[{Li et~al.(2024{\natexlab{b}})Li, Guo, Shao, Song, Bian, Zhang, and Wang}]{GNN-Ret}
Zijian Li, Qingyan Guo, Jiawei Shao, Lei Song, Jiang Bian, Jun Zhang, and Rui Wang. 2024{\natexlab{b}}.
\newblock Graph neural network enhanced retrieval for question answering of llms.
\newblock \emph{arXiv preprint arXiv:2406.06572}.

\bibitem[{Liang et~al.(2024)Liang, Niu, Zhang, Song, Wang, Yang, Xiong, Tang, Xi et~al.}]{PG-RAG}
Xun Liang, Simin Niu, Sensen Zhang, Shichao Song, Hanyu Wang, Jiawei Yang, Feiyu Xiong, Bo~Tang, Chenyang Xi, et~al. 2024.
\newblock Empowering large language models to set up a knowledge retrieval indexer via self-learning.
\newblock \emph{arXiv preprint arXiv:2405.16933}.

\bibitem[{Lin(2004)}]{lin2004rouge}
Chin-Yew Lin. 2004.
\newblock Rouge: A package for automatic evaluation of summaries.
\newblock In \emph{Text summarization branches out}, pages 74--81.

\bibitem[{Lin et~al.(2023)Lin, Asai, Li, Oguz, Lin, Mehdad, Yih, and Chen}]{lin2023train-dragon}
Sheng-Chieh Lin, Akari Asai, Minghan Li, Barlas Oguz, Jimmy Lin, Yashar Mehdad, Wen-tau Yih, and Xilun Chen. 2023.
\newblock How to train your dragon: Diverse augmentation towards generalizable dense retrieval.
\newblock \emph{arXiv preprint arXiv:2302.07452}.

\bibitem[{Liu et~al.(2024{\natexlab{a}})Liu, Zhang, Li, and Yao}]{EtD}
Guangyi Liu, Yongqi Zhang, Yong Li, and Quanming Yao. 2024{\natexlab{a}}.
\newblock Explore then determine: A gnn-llm synergy framework for reasoning over knowledge graph.
\newblock \emph{arXiv preprint arXiv:2406.01145}.

\bibitem[{Liu et~al.(2024{\natexlab{b}})Liu, Lin, Hewitt, Paranjape, Bevilacqua, Petroni, and Liang}]{liu2024lost-middle}
Nelson~F Liu, Kevin Lin, John Hewitt, Ashwin Paranjape, Michele Bevilacqua, Fabio Petroni, and Percy Liang. 2024{\natexlab{b}}.
\newblock Lost in the middle: How language models use long contexts.
\newblock \emph{Transactions of the Association for Computational Linguistics}, 12:157--173.

\bibitem[{Luo et~al.(2023)Luo, Li, Haffari, and Pan}]{RoG}
Linhao Luo, Yuan-Fang Li, Gholamreza Haffari, and Shirui Pan. 2023.
\newblock Reasoning on graphs: Faithful and interpretable large language model reasoning.
\newblock \emph{arXiv preprint arXiv:2310.01061}.

\bibitem[{Mavromatis and Karypis(2024)}]{GNN-RAG}
Costas Mavromatis and George Karypis. 2024.
\newblock Gnn-rag: Graph neural retrieval for large language model reasoning.
\newblock \emph{arXiv preprint arXiv:2405.20139}.

\bibitem[{Nogueira et~al.(2019)Nogueira, Lin, and Epistemic}]{nogueira2019doc2query}
Rodrigo Nogueira, Jimmy Lin, and AI~Epistemic. 2019.
\newblock From doc2query to doctttttquery.
\newblock \emph{Online preprint}, 6(2).

\bibitem[{Raffel et~al.(2020)Raffel, Shazeer, Roberts, Lee, Narang, Matena, Zhou, Li, and Liu}]{T5}
Colin Raffel, Noam Shazeer, Adam Roberts, Katherine Lee, Sharan Narang, Michael Matena, Yanqi Zhou, Wei Li, and Peter~J Liu. 2020.
\newblock Exploring the limits of transfer learning with a unified text-to-text transformer.
\newblock \emph{Journal of machine learning research}, 21(140):1--67.

\bibitem[{Ram et~al.(2023)Ram, Levine, Dalmedigos, Muhlgay, Shashua, Leyton-Brown, and Shoham}]{RALM}
Ori Ram, Yoav Levine, Itay Dalmedigos, Dor Muhlgay, Amnon Shashua, Kevin Leyton-Brown, and Yoav Shoham. 2023.
\newblock In-context retrieval-augmented language models.
\newblock \emph{Transactions of the Association for Computational Linguistics}, 11:1316--1331.

\bibitem[{Ramos et~al.(2003)}]{ramos2003using-tf-idf}
Juan Ramos et~al. 2003.
\newblock Using tf-idf to determine word relevance in document queries.
\newblock In \emph{Proceedings of the first instructional conference on machine learning}, volume 242, pages 29--48. Citeseer.

\bibitem[{Reimers and Gurevych(2019)}]{Sentence-BERT}
Nils Reimers and Iryna Gurevych. 2019.
\newblock \href {https://arxiv.org/abs/1908.10084} {Sentence-bert: Sentence embeddings using siamese bert-networks}.
\newblock In \emph{Proceedings of the 2019 Conference on Empirical Methods in Natural Language Processing}. Association for Computational Linguistics.

\bibitem[{Robertson et~al.(2009)Robertson, Zaragoza et~al.}]{robertson2009probabilistic-bm25}
Stephen Robertson, Hugo Zaragoza, et~al. 2009.
\newblock The probabilistic relevance framework: Bm25 and beyond.
\newblock \emph{Foundations and Trends{\textregistered} in Information Retrieval}, 3(4):333--389.

\bibitem[{Sun et~al.(2023)Sun, Xu, Tang, Wang, Lin, Gong, Shum, and Guo}]{ToG}
Jiashuo Sun, Chengjin Xu, Lumingyuan Tang, Saizhuo Wang, Chen Lin, Yeyun Gong, Heung-Yeung Shum, and Jian Guo. 2023.
\newblock Think-on-graph: Deep and responsible reasoning of large language model with knowledge graph.
\newblock \emph{arXiv preprint arXiv:2307.07697}.

\bibitem[{Team et~al.(2024)Team, Mesnard, Hardin, Dadashi, Bhupatiraju, Pathak, Sifre, Rivi{\`e}re, Kale, Love et~al.}]{team2024gemma}
Gemma Team, Thomas Mesnard, Cassidy Hardin, Robert Dadashi, Surya Bhupatiraju, Shreya Pathak, Laurent Sifre, Morgane Rivi{\`e}re, Mihir~Sanjay Kale, Juliette Love, et~al. 2024.
\newblock Gemma: Open models based on gemini research and technology.
\newblock \emph{arXiv preprint arXiv:2403.08295}.

\bibitem[{Touvron et~al.(2023)Touvron, Martin, Stone, Albert, Almahairi, Babaei, Bashlykov, Batra, Bhargava, Bhosale et~al.}]{touvron2023llama-2}
Hugo Touvron, Louis Martin, Kevin Stone, Peter Albert, Amjad Almahairi, Yasmine Babaei, Nikolay Bashlykov, Soumya Batra, Prajjwal Bhargava, Shruti Bhosale, et~al. 2023.
\newblock Llama 2: Open foundation and fine-tuned chat models.
\newblock \emph{arXiv preprint arXiv:2307.09288}.

\bibitem[{Trivedi et~al.(2022)Trivedi, Balasubramanian, Khot, and Sabharwal}]{trivedi2022interleaving-ralm}
Harsh Trivedi, Niranjan Balasubramanian, Tushar Khot, and Ashish Sabharwal. 2022.
\newblock Interleaving retrieval with chain-of-thought reasoning for knowledge-intensive multi-step questions.
\newblock \emph{arXiv preprint arXiv:2212.10509}.

\bibitem[{Tworkowski et~al.(2023)Tworkowski, Staniszewski, Pacek, Wu, Michalewski, and Miłoś}]{longllama}
Szymon Tworkowski, Konrad Staniszewski, Mikołaj Pacek, Yuhuai Wu, Henryk Michalewski, and Piotr Miłoś. 2023.
\newblock \href {https://arxiv.org/abs/2307.03170} {Focused transformer: Contrastive training for context scaling}.
\newblock \emph{Preprint}, arXiv:2307.03170.

\bibitem[{van~den Oord et~al.(2018)van~den Oord, Li, and Vinyals}]{CPC}
Aaron van~den Oord, Yazhe Li, and Oriol Vinyals. 2018.
\newblock Representation learning with contrastive predictive coding.
\newblock \emph{arXiv preprint arXiv:1807.03748v2}.

\bibitem[{Veli{\v{c}}kovi{\'c} et~al.(2017)Veli{\v{c}}kovi{\'c}, Cucurull, Casanova, Romero, Lio, and Bengio}]{GAT}
Petar Veli{\v{c}}kovi{\'c}, Guillem Cucurull, Arantxa Casanova, Adriana Romero, Pietro Lio, and Yoshua Bengio. 2017.
\newblock Graph attention networks.
\newblock \emph{arXiv preprint arXiv:1710.10903}.

\bibitem[{Wang et~al.(2022{\natexlab{a}})Wang, Pang, Chen, Phang, and Bowman}]{wang2022squality}
Alex Wang, Richard~Yuanzhe Pang, Angelica Chen, Jason Phang, and Samuel~R Bowman. 2022{\natexlab{a}}.
\newblock Squality: Building a long-document summarization dataset the hard way.
\newblock \emph{arXiv preprint arXiv:2205.11465}.

\bibitem[{Wang et~al.(2022{\natexlab{b}})Wang, Wei, Schuurmans, Le, Chi, Narang, Chowdhery, and Zhou}]{Self-consistency}
Xuezhi Wang, Jason Wei, Dale Schuurmans, Quoc Le, Ed~Chi, Sharan Narang, Aakanksha Chowdhery, and Denny Zhou. 2022{\natexlab{b}}.
\newblock Self-consistency improves chain of thought reasoning in language models.
\newblock \emph{arXiv preprint arXiv:2203.11171}.

\bibitem[{Wang et~al.(2024)Wang, Lipka, Rossi, Siu, Zhang, and Derr}]{KGP}
Yu~Wang, Nedim Lipka, Ryan~A Rossi, Alexa Siu, Ruiyi Zhang, and Tyler Derr. 2024.
\newblock Knowledge graph prompting for multi-document question answering.
\newblock In \emph{Proceedings of the AAAI Conference on Artificial Intelligence}, volume~38, pages 19206--19214.

\bibitem[{Wei et~al.(2022)Wei, Wang, Schuurmans, Bosma, Xia, Chi, Le, Zhou et~al.}]{wei2022few-shot-cot}
Jason Wei, Xuezhi Wang, Dale Schuurmans, Maarten Bosma, Fei Xia, Ed~Chi, Quoc~V Le, Denny Zhou, et~al. 2022.
\newblock Chain-of-thought prompting elicits reasoning in large language models.
\newblock \emph{Advances in neural information processing systems}, 35:24824--24837.

\bibitem[{Wu et~al.(2019)Wu, Souza, Zhang, Fifty, Yu, and Weinberger}]{SGC}
Felix Wu, Amauri Souza, Tianyi Zhang, Christopher Fifty, Tao Yu, and Kilian Weinberger. 2019.
\newblock Simplifying graph convolutional networks.
\newblock In \emph{International Conference on Machine Learning}, pages 6861--6871.

\bibitem[{Xu et~al.(2019)Xu, Hu, Leskovec, and Jegelka}]{GIN}
Keyulu Xu, Weihua Hu, Jure Leskovec, and Stefanie Jegelka. 2019.
\newblock How powerful are graph neural networks?
\newblock In \emph{International Conference on Learning Representations}.

\bibitem[{Yang et~al.(2024)Yang, Yu, Zhang, Cao, Xu, Zhang, Gonzalez, and Cui}]{BoT}
Ling Yang, Zhaochen Yu, Tianjun Zhang, Shiyi Cao, Minkai Xu, Wentao Zhang, Joseph~E Gonzalez, and Bin Cui. 2024.
\newblock Buffer of thoughts: Thought-augmented reasoning with large language models.
\newblock \emph{arXiv preprint arXiv:2406.04271}.

\bibitem[{Yao et~al.(2024)Yao, Yu, Zhao, Shafran, Griffiths, Cao, and Narasimhan}]{ToT}
Shunyu Yao, Dian Yu, Jeffrey Zhao, Izhak Shafran, Tom Griffiths, Yuan Cao, and Karthik Narasimhan. 2024.
\newblock Tree of thoughts: Deliberate problem solving with large language models.
\newblock \emph{Advances in Neural Information Processing Systems}, 36.

\bibitem[{Yu et~al.(2023)Yu, Xiong, Yu, and Liu}]{yu2023ralm}
Zichun Yu, Chenyan Xiong, Shi Yu, and Zhiyuan Liu. 2023.
\newblock Augmentation-adapted retriever improves generalization of language models as generic plug-in.
\newblock \emph{arXiv preprint arXiv:2305.17331}.

\bibitem[{Zhang et~al.(2019)Zhang, Kishore, Wu, Weinberger, and Artzi}]{zhang2019bertscore}
Tianyi Zhang, Varsha Kishore, Felix Wu, Kilian~Q Weinberger, and Yoav Artzi. 2019.
\newblock Bertscore: Evaluating text generation with bert.
\newblock \emph{arXiv preprint arXiv:1904.09675}.

\bibitem[{Zheng et~al.(2023)Zheng, Mishra, Chen, Cheng, Chi, Le, and Zhou}]{Takeastepback}
Huaixiu~Steven Zheng, Swaroop Mishra, Xinyun Chen, Heng-Tze Cheng, Ed~H Chi, Quoc~V Le, and Denny Zhou. 2023.
\newblock Take a step back: Evoking reasoning via abstraction in large language models.
\newblock \emph{arXiv preprint arXiv:2310.06117}.

\bibitem[{Zhong et~al.(2021)Zhong, Yin, Yu, Zaidi, Mutuma, Jha, Awadallah, Celikyilmaz, Liu, Qiu et~al.}]{QMSum}
Ming Zhong, Da~Yin, Tao Yu, Ahmad Zaidi, Mutethia Mutuma, Rahul Jha, Ahmed~Hassan Awadallah, Asli Celikyilmaz, Yang Liu, Xipeng Qiu, et~al. 2021.
\newblock Qmsum: A new benchmark for query-based multi-domain meeting summarization.
\newblock \emph{arXiv preprint arXiv:2104.05938}.

\bibitem[{Zhu and Ghahramani(2002)}]{LabelPropagation}
Xiaojin Zhu and Zoubin Ghahramani. 2002.
\newblock Learning from labeled and unlabeled data with label propagation.
\newblock \emph{ProQuest number: information to all users}.

\end{thebibliography}

\clearpage
\appendix
% \onecolumn

\section{Experimental Details}
\label{appendix:exp_details}

\subsection{Dataset}

We present dataset statistics in Table~\ref{tab:dataset_sta}. 
Due to the limited budget, we randomly select training and test samples for the training and test set and calculate the average input and output token lengths using the LLaMA-2 tokenizer~\citep{touvron2023llama-2} (samples with short input lengths are filtered out). 

We evaluate our proposed method on four long-context summarization datasets, \textit{i.e.}, AcademicEval~\citep{Thought-Retriever}, QMSum~\citep{QMSum}, WCEP~\citep{WCEP}, and BookSum~\citep{kryscinski2021booksum}.

\begin{itemize}
\item \textbf{QMSum}~\citep{QMSum}. QMSum is a query-based summarization dataset that features lengthy meeting transcripts, specific queries, and general queries. Specific queries focus on query-based summarization, and general queries are questions that summarize the entire meeting transcript, such as “Summarize the whole meeting.”
We \textbf{only use} ``\textit{general queries}” for evaluating global summarization. 
\item \textbf{AcademicEval}~\citep{Thought-Retriever}. AcademicEval collects scientific papers from arXiv for abstract and related work writing. We use the abstract writing subset, which provides the main body of a paper as input and generates the predicted abstract. 
\item \textbf{WCEP}~\citep{WCEP}. WCEP is a multi-document summarization dataset about news events, which requires comprehensive consideration of the contents of multiple documents. 
\item \textbf{BookSum}~\citep{kryscinski2021booksum}. BookSum features long-form narrative summarization, which covers source documents from the literature domain and includes highly abstractive human-written summaries. 
\end{itemize}

\begin{table*}[h]
  % \footnotesize
  \centering
\begin{center}
  \begin{tabular}{ccccc}
    \toprule
    Dataset & \#Train & \#Test & \makecell{Average Input \\ Token Length} & \makecell{Average Output \\ Token Length} \\
    \midrule
    % Results
    QMSum~\citep{QMSum}
    & 162 & 30 & 17K & 0.1K \\
    AcademicEval~\citep{Thought-Retriever}
    & 400 & 30 & 13K & 0.3K \\
    WCEP~\citep{WCEP}
    & 400 & 30 & 11K & 0.05K \\
    BookSum~\citep{kryscinski2021booksum}
    & 400 & 30 & 16K & 1K \\
    % Results
    \bottomrule
  \end{tabular}
\end{center}
\caption{Dataset statistics}
  \label{tab:dataset_sta}
\end{table*}

\subsection{Baselines}

We present detailed descriptions of adopted baselines.
\begin{itemize}
\item \textbf{Node2Vec}~\citep{grover2016node2vec}. Node2Vec generates node embeddings for graphs by simulating biased random walks to capture both local and global structural properties of nodes.
\item \textbf{BM25}~\citep{robertson2009probabilistic-bm25}, \textbf{TF-IDF}~\citep{ramos2003using-tf-idf}. BM25 ranks documents based on term frequency, inverse document frequency, and document length normalization, while TF-IDF evaluates the importance of a term in a document relative to a corpus by combining term frequency and inverse document frequency.
\item \textbf{Contriever}~\citep{izacard2021unsupervised-contriever}, \textbf{DPR}~\citep{karpukhin2020dense-dpr}, \textbf{Dragon}~\citep{lin2023train-dragon}, \textbf{SBERT}~\citep{Sentence-BERT}. Contriever is a self-supervised dense retriever that learns unsupervised document embeddings for information retrieval, DPR (Dense Passage Retriever) is a bi-encoder model that retrieves relevant passages by training on question-passage pairs, Dragon is a dense retrieval model optimized through diverse augmentation for generalizable dense retrieval, and SBERT (Sentence-BERT) is a modification of BERT that generates semantically meaningful sentence embeddings for tasks like similarity and clustering using a siamese network structure.
\item \textbf{BM25+DPR}. BM25+DPR with Reciprocal Rerank Fusion is a hybrid retrieval method that combines the strengths of BM25's lexical matching and DPR's dense embeddings by reranking results from both models using a reciprocal rank fusion strategy to improve retrieval accuracy.
\item \textbf{Gemma-8K}~\citep{team2024gemma}, \textbf{Mistral-8K}~\citep{jiang2023mistral}. Gemma-8K and Mistral-8K are LLMs with relatively long context window lengths. 
\item \textbf{Full Context}. We feed all inputs to LLMs for summary generation. If the input length exceeds the context window limit, we randomly sample continuous text spans of maximum length multiple times to feed into LLMs and calculate the average result.
\item \textbf{Thought-R}~\citep{Thought-Retriever}. Thought Retriever (Thought-R) generates thoughts for a series of simulated queries and appends them to the retrieval corpus as high-level knowledge. 
\end{itemize}

\subsection{Additional Explanation on Training Objective}

Given a graph that consists of document chunks and response nodes, we expect that the learned node embeddings $\mathbf{h}_{v}^{(L)}$ can adaptively reflect the semantic similarity to a given query $\mathbf{q}$. 
In other words, we expect that we can select the node $\mathbf{v}$ with the largest semantic similarity to $\mathbf{q}$ according to the formula $\operatorname{Sim}(\mathbf{q}, \mathbf{v}) = \mathbf{E}_{\mathbf{q}}(\mathbf{q})^{T} \cdot \mathbf{h}_{v}^{(L)}$. 
To this end, we need to find out which node has the highest semantic similarity with $\mathbf{q}$ and use this as a supervision signal for model optimization. 
Therefore, we utilize BERTScore~\citep{zhang2019bertscore} to obtain a node ranking list $\mathbf{M}_{i}$, which exactly serves as supervision signals. 
\begin{equation}
\mathbf{M}_{i} = [\mathbf{h}_{+}^{(L)}, \mathbf{h}_{1}^{(L)}, \cdots, \mathbf{h}_{|\mathbf{C}| + |\mathbf{T}|}^{(L)}]
\end{equation}

For contrastive loss in Equation~\ref{formula:cl}, we regard $\mathbf{h}_{+}^{(L)}$ as the positive and $[\mathbf{h}_{1}^{(L)}, \cdots, \mathbf{h}_{|\mathbf{C}| + |\mathbf{T}|}^{(L)}]$ as the negatives to conduct contrastive learning~\citep{CPC}. For a given query $\mathbf{q}$, the contrastive learning objective will bring $\mathbf{E}_{\mathbf{q}}(\mathbf{q})$ and $\mathbf{h}_{+}^{(L)}$ closer in the semantic embedding space while increasing the distance between $\mathbf{E}_{\mathbf{q}}(\mathbf{q})$ and $[\mathbf{h}_{1}^{(L)}, \cdots, \mathbf{h}_{|\mathbf{C}| + |\mathbf{T}|}^{(L)}]$ (the symbols are simplified here for convenience of description).

Similarly, in Equation~\ref{formula:cl_aux}, we expect to impose a stricter constraint on the learned node embedding based on the node ranking list $\mathbf{M}_{i}$. 
Although Equation~\ref{formula:cl} shortens the semantic distance between $\mathbf{E}_{\mathbf{q}}(\mathbf{q})$ and $\mathbf{h}_{+}^{(L)}$, it does not take into account the relative ranking between negative samples. 
For example, the semantic similarity between $\mathbf{E}_{\mathbf{q}}(\mathbf{q})$ and $\mathbf{h}_{1}^{(L)}$ is higher than that between $\mathbf{h}_{2}^{(L)}$. 
Formally, given $\mathbf{h}_{i}^{(L)}, \mathbf{h}_{j}^{(L)} \in \mathbf{M}_{i}$ that satisfies $\operatorname{rank}(\mathbf{h}_{j}^{(L)}) > \operatorname{rank}(\mathbf{h}_{i}^{(L)})$, Equation~\ref{formula:cl_aux} will explicitly optimize in the direction of ${\mathbf{E}_{\mathbf{q}}(\mathbf{q})}^{\top} \mathbf{h}_{j}^{(L)} < {\mathbf{E}_{\mathbf{q}}(\mathbf{q})}^{\top} \mathbf{h}_{i}^{(L)}$, thus imposing stricter constraints to the pair-wise ranking.

\subsection{Additional Implementation Details}

In the stage of graph construction, due to the number and randomness of the simulated queries, there may be some isolated nodes, and we just keep them in the graph with self-loop edges. During model optimization, BERTScore is pre-computed for efficient training. 

In the training stage, we use the Adam optimizer for model training and gradually decay the learning rate from 1e-3 to 0 with the LambdaLR scheduler. 
We present detailed hyper-parameters on QMSum, AcademicEval, WCEP, and BookSum datasets in Table~\ref{tab:hyper-parameters}. 
We implement our proposed method using PyTorch and Deep Graph Library (DGL), and all the experiments are conducted on a single RTX 3080 GPU. 
As for LLMs, we rely on API calling from Together AI\footnote{\space https://www.together.ai/} to obtain responses.

For metrics, we adopt Rouge-1 (R-1), Rouge-2 (R-2), and Rouge-L  (R-L)~\citep{lin2004rouge} to assess the text alignment between the reference summaries and the predicted content generated by our proposed method. 
If a global summarization query has multiple reference summaries, we calculate the Rouge-L/1/2 of the predicted summary and all references, respectively, and take the maximum value as the final evaluation result. We follow this setting in all experiments, including the baseline evaluation.

\begin{table*}[h]
  % \footnotesize
  \centering
\begin{center}
  \begin{tabular}{ccccc}
    \toprule
    Datasets & \makecell{QMSum} & \makecell{AcademicEval} & \makecell{WCEP} & \makecell{BookSum} \\
    \midrule
    % Results
    \#GAT Layers
    & 2 & 2 & 2 & 2 \\
    \#GAT Heads
    & 4 & 4 & 4 & 4 \\
    Batch Size
    & 32 & 32 & 32 & 32 \\
    Epoch
    & 150 & 150 & 150 & 150 \\
    Learning Rate
    & 1e-3 & 1e-3 & 1e-3 & 1e-3 \\
    Hidden Dimension
    & 768 & 768 & 768 & 768 \\
    Dropout Rate
    & 0.2 & 0.0 & 0.1 & 0.2 \\
    Loss Coefficient $\alpha$
    & 0.9 & 0.6 & 0.7 & 0.2 \\
    % Results
    \bottomrule
  \end{tabular}
\end{center}
\caption{Hyper-parameters}
  \label{tab:hyper-parameters}
\end{table*}

\subsection{More Comparison Experiments}

We conduct extensive experiments on GovReport~\citep{GovReport} and SQuALITY~\citep{wang2022squality} datasets, and the results are shown in Table~\ref{tab:more_results}, which demonstrate our proposed GoR is still competitive among baselines on these two datasets. 

\begin{table*}[h]
  \footnotesize
  \centering
  \begin{center}
  \begin{tabular}{ccccccc}
    \toprule
    \multirow{2}{*}{Model} & \multicolumn{3}{c}{GovReport} & \multicolumn{3}{c}{SQuALITY} \\
    \cmidrule(r){2-4} \cmidrule(r){5-7} 
    \multicolumn{1}{c}{} & R-L & R-1 & R-2 & R-L & R-1 & R-2 \\
    \midrule
    % Results
    Node2Vec
    & 18.1 & 36.7 & 12.4 
    & 17.0 & 32.9 & 7.7 \\
    \midrule
    BM25
    & 18.2 & 39.2 & 13.0 
    & 17.0 & 31.4 & 8.1 \\
    TF-IDF
    & 18.1 & 39.2 & 12.8 
    & 17.0 & 31.4 & 8.1 \\
    \midrule
    Contriever
    & 20.2 & 39.8 & \textbf{17.6} 
    & 16.8 & 32.6 & 8.3 \\
    DPR
    & 19.1 & 39.4 & 15.5 
    & 17.4 & 33.1 & 8.4 \\
    Dragon
    & 19.6 & 38.2 & 16.0 
    & 16.2 & 29.6 & 7.5 \\
    SBERT
    & 20.0 & 39.8 & 15.8 
    & 17.1 & 32.1 & 7.8 \\
    \midrule
    BM25+DPR
    & 19.4 & 37.4 & 15.0 
    & 16.6 & 31.5 & 7.4 \\
    \midrule
    Gemma-8K
    & 17.4 & 33.8 & 11.4 
    & 12.9 & 19.7 & 5.8 \\
    Mistral-8K
    & 16.0 & 28.9 & 9.4 
    & 16.9 & 32.2 & 8.1 \\
    \midrule
    Full Context
    & 18.4 & 39.1 & 13.8 
    & \textbf{17.8} & \textbf{34.0} & \textbf{8.8} \\
    \midrule
    Thought-R
    & 20.4 & 40.3 & 17.0 
    & 17.3 & 32.0 & 8.0 \\
    \midrule
    \textbf{GoR (Ours)}
    & \textbf{20.9} & \textbf{41.4} & 16.8 
    & \textbf{17.8} & \textbf{34.0} & 8.5 \\
    % Results
    \bottomrule
  \end{tabular}
  \end{center}
  \caption{\textbf{Experimental results on GovReport and SQuALITY datasets over long-context global summarization tasks w.r.t. Rouge-L (R-L), Rouge-1 (R-1), and Rouge-2 (R-2)}. Note that the average LLM input token length of GoR and retriever-based baselines is $6 \times 256$, which is about 1.5K. (\textbf{BOLD} indicates the best score)}
    \label{tab:more_results}
\end{table*}

\subsection{More Ablation Experiments}

We conduct extensive ablation experiments on QMSum~\citep{QMSum} and AcademicEval~\citep{Thought-Retriever} datasets, and the results are shown in Table~\ref{tab:add_ablation}. 

\begin{table}[h]
  \footnotesize
    \centering
  \begin{center}
  \begin{tabular}{@{}ccccccc}
    \toprule
    \multirow{2}{*}{Variant} & \multicolumn{3}{c}{QMSum} & \multicolumn{3}{c}{AcademicEval} \\
    \cmidrule(r){2-4} \cmidrule(r){5-7}
    \multicolumn{1}{c}{} & R-L & R-1 & R-2 & R-L & R-1 & R-2 \\
    \midrule
    % Results
    w/o train
    & 18.2 & 33.0 & 7.6 
    & 23.3 & 45.0 & 15.5 \\
    w/o $\mathcal{L}_{\mathrm{CL}}$
    & 18.4 & 33.3 & 6.9 
    & 23.5 & 44.9 & 15.5 \\
    w/o $\mathcal{L}_{\mathrm{RANK}}$
    & 19.6 & 33.1 & \textbf{7.8} 
    & 23.1 & 44.4 & 15.1 \\
    w/o in-b neg
    & \textbf{19.8} & \textbf{34.7} & \textbf{7.8} 
    & 24.5 & 46.4 & 16.5 \\
    w/ sup
    & 18.1 & 32.3 & 6.9 
    & 21.4 & 43.3 & 13.9 \\
    \midrule
    \textbf{GoR}
    & \textbf{19.8} & 34.5 & \textbf{7.8} 
    & \textbf{24.7} & \textbf{46.5} & \textbf{17.3} \\
    % Results
    \bottomrule
  \end{tabular}
  \end{center}
  \caption{\textbf{Ablation study on QMSum and AcademicEval datasets w.r.t. R-L, R-1, and R-2}.}
    \label{tab:add_ablation}
\end{table}

\subsection{Impact of GNN Architectures}
\begin{figure*}[t]
\begin{center}
    \centering
    \begin{minipage}{0.49\linewidth}
        \includegraphics[width=\linewidth]{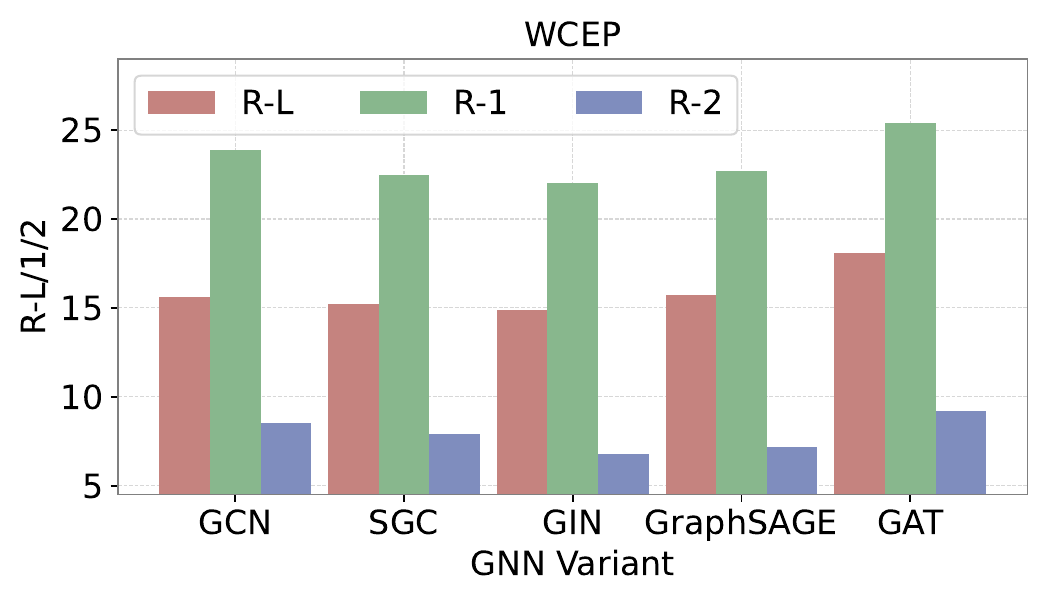}
        % \label{fig:gnn_wcep}
    \end{minipage}
    \hfill
    \begin{minipage}{0.49\linewidth}
        \includegraphics[width=\linewidth]{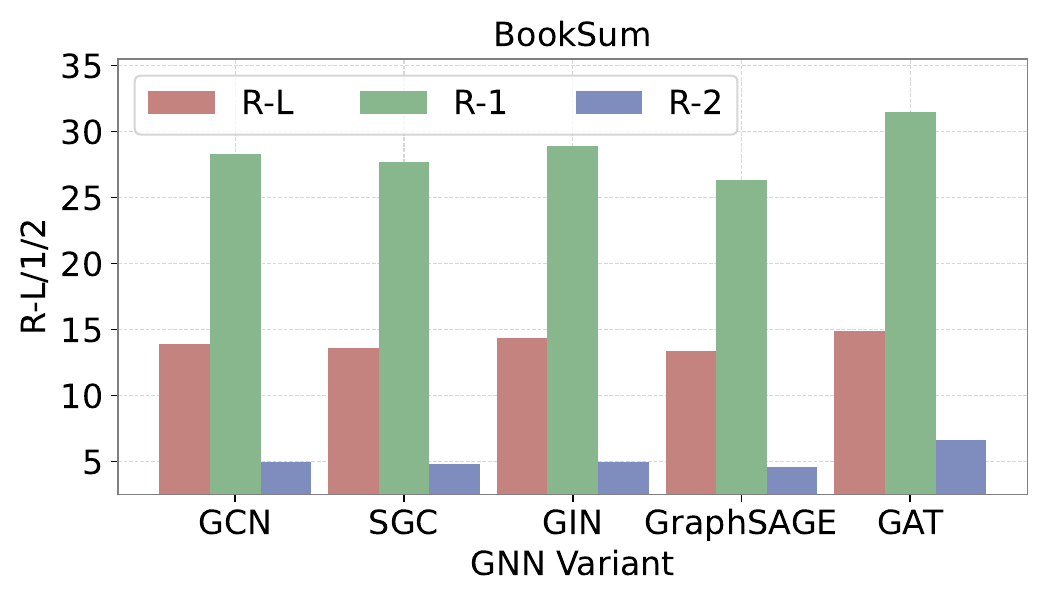}
        % \label{fig:gnn_book}
    \end{minipage}
    \caption{\textbf{Impact of GNN architectures w.r.t. R-L, R-1, and R-2}. The left figure shows results on the WCEP dataset, while the right one shows results with the BookSum dataset.}
    \label{fig:gnn_arc}
\end{center}
\end{figure*}

GNNs play a vital role in learning node embeddings. we explore various GNN architectures to study their impact on learning node embeddings, including GCN~\citep{GCN}, SGC~\citep{SGC}, GIN~\citep{GIN}, and GraphSAGE~\citep{GraphSAGE}. Our findings, illustrated in Figure~\ref{fig:gnn_arc}, show that GAT outperforms the other architectures. This is because GAT considers the significance of neighboring nodes when updating node embeddings, allowing the model to effectively capture essential information from the nodes. 
Among the other architectures, GraphSAGE performs poorly due to its unstable neighbor sampling mechanism. 

Overall, GAT reaches the best results, which shows that considering the importance of neighboring nodes is effective in mining complicated correlations and is critical to improving performance.

\subsection{Impact of the Number of Simulated Queries During Training}
\begin{figure}[h]
\begin{center}
    \centering
    \begin{subfigure}{0.49\linewidth}
        \centering
        \includegraphics[width=1.0\linewidth]{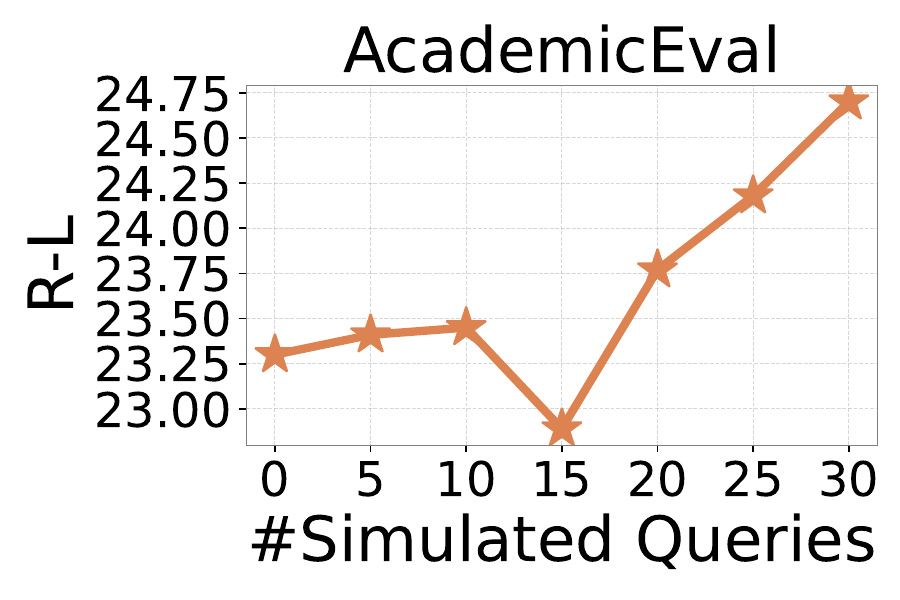}
        \label{fig:query_aca}
    \end{subfigure}
    \centering
    \begin{subfigure}{0.49\linewidth}
        \centering
        \includegraphics[width=1.0\linewidth]{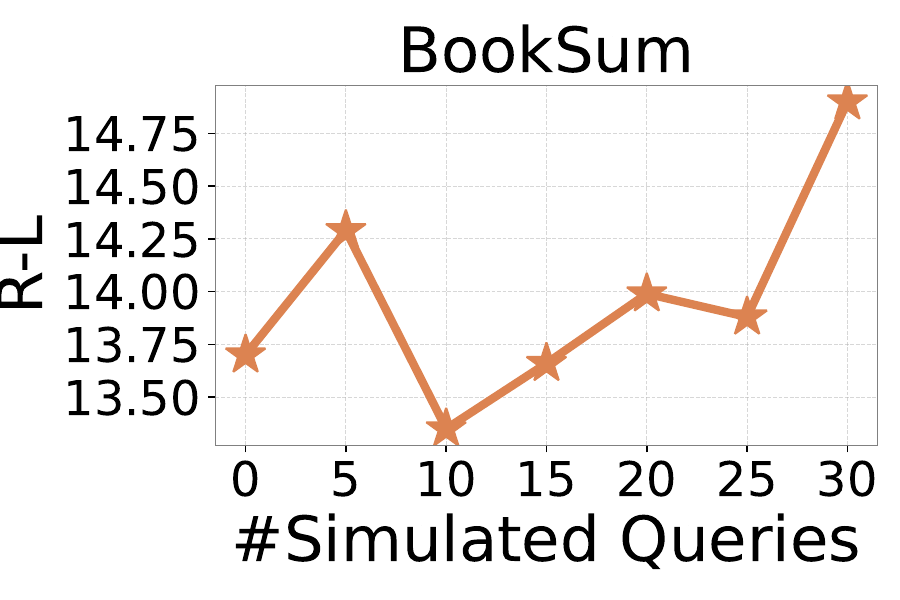}
        \label{fig:query_book}
    \end{subfigure}
    \caption{\textbf{Impact of the number of simulated queries during training w.r.t. R-L}. We show the additional results on the AcademicEval and BookSum datasets.}
    \label{fig:add_query}
\end{center}
\end{figure}

We show additional results on the AcademicEval and BookSum datasets in Figure ~\ref{fig:add_query}. 

\subsection{Supervised Training on Global Summarization Queries}
\begin{figure}[h]
\begin{center}
    \centering
    \begin{minipage}{0.49\linewidth}
        \includegraphics[width=\linewidth]{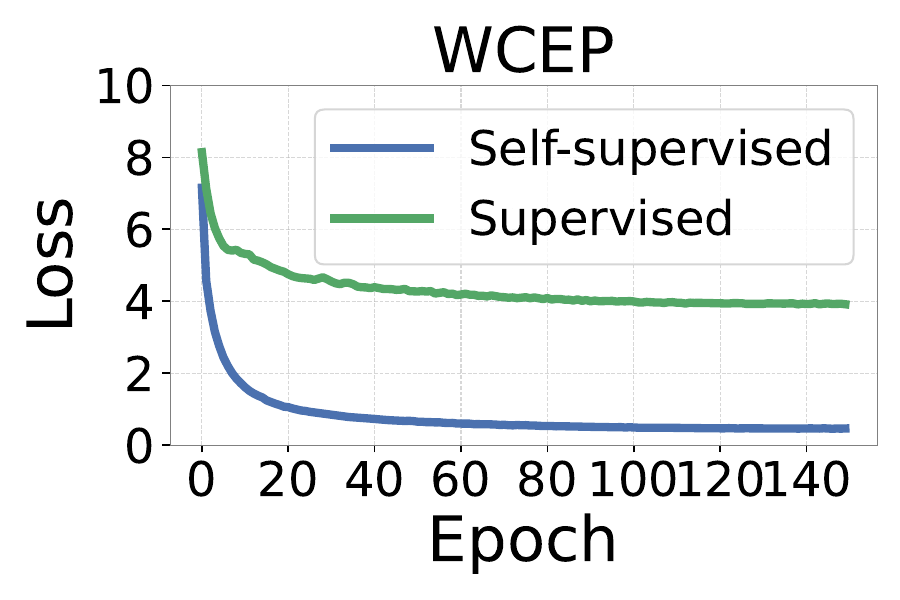}
        \label{fig:loss_wcep}
    \end{minipage}
    \hfill
    \begin{minipage}{0.49\linewidth}
        \includegraphics[width=\linewidth]{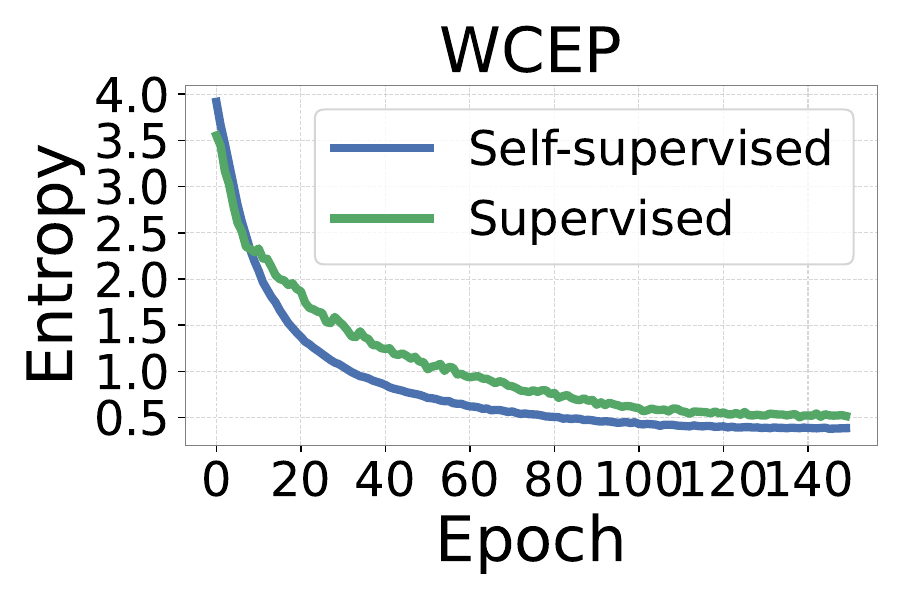}
        \label{fig:entropy_wcep}
    \end{minipage}
    \caption{\textbf{Differences between self-supervised and supervised training w.r.t. loss and entropy}. We show the loss and entropy curve during training on the WCEP dataset.}
    \label{fig:add_loss_entropy}
\end{center}
\end{figure}

We show additional results on the WCEP dataset in Figure ~\ref{fig:add_loss_entropy}.

\section{Additional Related Work}
\label{appendix:add_related_work}

\textbf{Historical Response Utilization of LLMs.} 
Little work has been done on this under-explored topic. 
Thought-retriever~\citep{Thought-Retriever} saves the historical responses of each user-LLM interaction as high-level and informative thoughts to expand the retrieval corpus for future user queries. 
However, the intricate correlations among thoughts are neglected, leaving room for further improvement. 

Another line of work is the Chain-of-Thought (CoT), which is similar to our approach in terms of utilizing LLM historical responses and has been regarded as an effective means to enhance the reasoning ability of LLM during inference time in recent years. 
Few-shot CoT~\citep{wei2022few-shot-cot} and zero-shot CoT~\citep{kojima2022zero-shot-cot} elicit intermediate reasoning paths by prompting LLMs with several demonstrations or just appending \textit{"Let's think step by step."} 
Self-consistency~\citep{Self-consistency} samples diverse reasoning paths and conducts a majority vote to obtain the final answer. 
ToT~\citep{ToT} and GoT~\citep{GoT} take a further step by integrating a tree or graph structure to manage its historical reasoning paths, enabling more flexible reasoning and reflection of LLMs. 
By memorizing the solution paradigms of various queries into different templates, BoT~\citep{BoT} pushes LLM's reasoning ability to a new level. 
Although the above-mentioned CoT-series approaches improve the reasoning capabilities of LLMs by utilizing chains of intermediate reasoning responses, they usually concentrate on one specific QA-type query and cannot generalize to and benefit other queries. 
Moreover, only a small number of historical reasoning responses are retained for the final generation, while most of the rest are just discarded.

\clearpage
\section{Case Study}
\label{appendix:case_study}

In this section, we provide a case study of GoR and other baseline methods' summarization.

\begin{center}
\begin{tcolorbox}[title={The Summary Generated by GoR}]
{
An asteroid called 1999 KW4 will make a close approach to Earth this weekend, with its own small moon in tow. The asteroid is estimated to be around 1.5 km wide and will be roughly 3.2 million miles from Earth at its closest point. NASA astronomers are interested in studying the asteroid due to its close proximity and the fact that it is a binary system, meaning it has its own moon. The last time the asteroid will make a close approach to Earth will be in 2036. Additionally, a recent study using data from NASA's Kepler Space Telescope suggests that comets may have delivered water to Earth, as the ratio of two types of water molecules on comets matches that in Earth's oceans. The new algorithm used in the study is more sensitive to small planets the size of Earth and could help in the search for Earth-like planets.
}
\label{appendix:p3}
\end{tcolorbox}
\end{center}

\begin{center}
\begin{tcolorbox}[title={The Summary Generated by Contriever}]
{
Asteroid 2019 JH7 recently flew past Earth, and NASA observed that the asteroid's trajectory falls under the "Earth Close Approach" category. The observations made by NASA's Jet Propulsion Laboratory (JPL) in Pasadena, California, suggest that comets could have delivered water to Earth. The study found that the ratio of two types of water molecules on the comet matches that in the Earth's oceans, which could imply that comets delivered a large fraction of water to the outer reaches of the solar system. The observations made using data from the SOFIA telescope suggest that all comets could have a heavy-to-regular water ratio similar to Earth's oceans, which could imply that comets delivered some water to Earth. Previously, measuring this ratio was difficult, and ground and space telescopes could only study this level of detail in comets when they pass near Earth.
}
\label{appendix:p4}
\end{tcolorbox}
\end{center}

\begin{center}
\begin{tcolorbox}[title={Reference Summary}]
{
Binary Aten asteroid (66391) 1999 KW4 and its minor-planet moon make their closest-ever recorded flyby of Earth at 3.2 million miles away. The asteroid will approach even closer at 0.0155 AU (2,320,000 km) from Earth in 2036, and is the largest asteroid to approach Earth until (4953) 1990 MU in June 2027.
}
\label{appendix:p5}
\end{tcolorbox}
\end{center}

From the above example, we can draw conclusions. 
(1) GoR summarizes several keywords that appear in the reference summary, such as ``1999 KW4" and ``3.2 million miles", etc., but Contriever fails to extract this crucial information.
(2) From a global perspective, the summary generated by GoR is more relevant and consistent with the reference summary. However, the summary generated by Contriever focuses too much on local details and ignores the main idea of the original article.

\clearpage
\section{LLM Prompts}
\label{appendix:prompts}

In this section, we present LLM prompts used in GoR, including user query simulation, RAG, and LLM-as-a-Judge prompts.

\subsection{LLM Prompts for User Query Simulation}

\begin{center}
\begin{tcolorbox}[title={Prompt for User Query Simulation}]
{
You are a great questioner of any text, and are adept at asking valuable and insightful questions. Your goal is to generate 1 summary question for the text provided below. The generated summary question should try to simulate the tone of human questions as much as possible, and make sure that the generated question must be interrogative sentences and a summary question. Important! Please make sure this text must be a complete and non-redundant answer to the generated summary question. Please directly output the generated summary question, do not output irrelevant text. \\

DOCUMENT: \\
\{document\}
}
\label{appendix:p1}
\end{tcolorbox}
\end{center}

\subsection{LLM Prompts for RAG}

\begin{center}
\begin{tcolorbox}[title={RAG Prompt}]
{
Refer to the following supporting materials and answer the question with brief but complete explanations. \\

SUPPORTING MATERIALS: \\
\{materials\} \\

QUESTION: \\
\{question\}
}
\label{appendix:p2}
\end{tcolorbox}
\end{center}

\subsection{LLM Prompts for LLM-as-a-Judge}

We construct LLM-as-a-Judge prompts following~\cite{GraphRAG} and~\cite{guo2024lightrag} with some minor changes.

\begin{center}
\begin{tcolorbox}[title={LLM-as-a-Judge Prompt - Instruction}]
{
---Role--- \\
You are an expert tasked with evaluating two answers to the same question based on four criteria: Comprehensiveness, Diversity, and Empowerment. \\

---Goal--- \\
You will evaluate two answers to the same question based on four criteria: Comprehensiveness, Diversity, and Empowerment. \\

Comprehensiveness: How much detail does the answer provide to cover all aspects and details of the question? \\
Diversity: How varied and rich is the answer in providing different perspectives and insights on the question? \\
Empowerment: How well does the answer help the reader understand and make informed judgments about the topic? \\

For each criterion, choose the better answer (either Answer 1 or Answer 2) and explain why. Then, select an overall winner based on these three categories. \\
}
\label{appendix:p33}
\end{tcolorbox}
\end{center}

\begin{center}
\begin{tcolorbox}[title={LLM-as-a-Judge Prompt - Input}]
{
Here is the question: \{query\} \\

Here are the two answers: Answer 1: \{answer1\}; Answer 2: \{answer2\} \\

Evaluate both answers using the three criteria listed above and provide detailed explanations for each criterion. \\

Avoiding any potential bias and ensuring that the order in which the answers were presented does not affect your judgment. \\

Output your evaluation in the following JSON format: \\

\{\{
    "Comprehensiveness": \{\{ "Winner": "[Answer 1 or Answer 2]", "Explanation": "[Provide explanation here]" \}\}, \\
    "Diversity": \{\{ "Winner": "[Answer 1 or Answer 2]", "Explanation": "[Provide explanation here]" \}\}, \\
    "Empowerment": \{\{ "Winner": "[Answer 1 or Answer 2]", "Explanation": "[Provide explanation here]" \}\}, \\
    "Overall Winner": \{\{"Winner": "[Answer 1 or Answer 2]", "Explanation": "[Summarize why this answer is the overall winner based on the three criteria]" \}\}
\}\}
}
\label{appendix:p44}
\end{tcolorbox}
\end{center}

\newpage
\section{Broader Impacts}

In the era of LLMs, countless interactions take place between users and these models on a daily basis, resulting in the generation of a vast amount of historical responses. Our proposed method demonstrates that these historical responses hold significant potential and can be effectively leveraged to further improve the quality of future responses generated by LLMs. By analyzing and reusing these past outputs, we can not only refine and enhance the overall performance of the models but also reduce computational overhead. This approach highlights the untapped value of historical data in optimizing response generation while making the process more efficient and resource-friendly.

\end{document}